\documentclass[journal]{IEEEtran}

\usepackage{amsmath,epsfig,lipsum}
\usepackage{subfigure}
\usepackage{multirow}
\usepackage{booktabs}
\usepackage{makecell}
\usepackage{cite}
\usepackage{color}
\usepackage{algorithmic}
\usepackage{graphicx}
\usepackage{textcomp}
\usepackage{xcolor}
\usepackage{placeins}
\usepackage{float}
\usepackage{tabularx,colortbl}
\usepackage{graphicx}
\usepackage{epstopdf}
\usepackage{cuted}
\usepackage[none]{hyphenat}

\ifCLASSINFOpdf
\else
\fi

\begin{document}\sloppy{}
	\title{Spatio-temporal Correlation Guided Geometric Partitioning for Versatile Video Coding}
	\author{Xuewei~Meng, 
		Chuanmin~Jia,
		Xinfeng~Zhang,~\IEEEmembership{Senior~Member,~IEEE,}
		Shanshe~Wang,~\IEEEmembership{Member,~IEEE,}
		and~Siwei~Ma,~\IEEEmembership{Senior~Member,~IEEE}

	\thanks{
		This work was supported in part by the National Natural Science Foundation of China (62025101, 62101007, U20A20184), National Key Research and Development Project (2019YFF0302703), National Postdoctoral Program for Innovative Talents (BX2021009), China Postdoctoral Science Foundation (2020M680238), PKU-Baidu Fund (2019BD003), and also supported by High performance Computing Platform of Peking University, which are gratefully acknowledged. The associate editor coordinating the review of this manuscript and approving it for publication was Dr. S{\'e}rgio de Faria. (Corresponding author: Chuanmin Jia.)
		
		Xuewei Meng, Chuanmin Jia, Shanshe Wang and Siwei Ma are with the Institute of Digital Media, Department of Electronics Engineering and Computer Science, Peking University, Beijing 100871, China~(e-mail: \{xwmeng, cmjia, sswang, swma\}@pku.edu.cn).
		
		Shanshe Wang and Siwei Ma are also with Information Technology R\&D Innovation Center of Peking University, Shaoxing 312000, China, and Peng Cheng Laboratory, Shenzhen 518066, China.
		
		Xinfeng Zhang is with the School of Computer Science and Technology, University of Chinese Academy of Sciences, Beijing 100190, China~(e-mail: xfzhang@ucas.ac.cn).}
    }

	\markboth{Journal of \LaTeX\ Class Files,~Vol.~XX, No.~X, November~2021}%
	{Shell \MakeLowercase{\textit{et al.}}: Bare Demo of IEEEtran.cls for IEEE Journals}
	\maketitle
	\begin{abstract}
		Geometric partitioning has attracted increasing attention by its remarkable motion field description capability in the hybrid video coding framework. However, the existing geometric partitioning~(GEO) scheme in Versatile Video Coding~(VVC) causes a non-negligible burden for signaling the side information. Consequently, the coding efficiency is limited. In view of this, we propose a spatio-temporal correlation guided geometric partitioning~(STGEO) scheme to efficiently describe the object information in the motion field of video coding. The proposed method can economize the bits consumed for side information signaling, including the partitioning mode and motion information. We firstly analyze the characteristics of partitioning mode decision and motion vector selection in a statistically-sound way. Based on the observed spatio-temporal correlation, we design a mode prediction and coding method to reduce the overhead for representing the above mentioned side information. The main idea is to predict the STGEO modes and motion candidates that have higher selection possibilities, which can guide the entropy coding, i.e., representing the predicted high-probability modes and motion candidates with fewer bits. In particular, the high-probability STGEO modes are predicted based on the edge information and history modes of adjacent STGEO-coded blocks. The corresponding motion information is represented by the index in a merge candidate list, which is adaptively inferred based on the off-line trained merge candidate selection probability. Simulation results show that the proposed approach achieves 0.95\% and 1.98\% bit-rate savings on average compared to VTM-8.0 without GEO for Random Access and Low-Delay B configurations, respectively.
	\end{abstract}
	
	
	\IEEEpeerreviewmaketitle
	
	\makeatletter
	\newcommand{\thickhline}{
		\noalign {\ifnum 0=`}\fi \hrule height 1pt
		\futurelet \reserved@a \@xhline
	}
	
	\section{Introduction}
	\IEEEPARstart{T}{he} hybrid video coding standards like Advanced Video Coding~(AVC)~\cite{AVC} and High Efficiency Video Coding~(HEVC)~\cite{HEVC} show remarkable compression capabilities on digital video. To meet the increasing requirements for high-resolution videos with limited transmission and memory capacity, a new video coding standard, Versatile Video Coding~(VVC)~\cite{VVC}, was developed by the Joint Video Experts Team~(JVET)~\cite{CFP} since October 2017. For these video coding standards, motion compensation is one of the fundamental methods of achieving high compression ratios. Generally, block-based motion compensation is widely used. In particular, the motion information is first derived by motion estimation~(ME) for each non-overlapped blocks at the encoder. Then the current frame is predicted from previously coded frames by block-based motion compensation using the motion information.
	
	Rectangular block partitioning is widely used for block-based motion compensation in video coding standards, such as $16 \times 16$ macroblock based partitioning in AVC, quadtree-based~(QT) block partitioning structure~\cite{kim2012block} and non-square QT structure~\cite{NSQT} in HEVC, and the quadtree with a nested multi-type tree~(QT-MTT) using binary tree and ternary tree partitioning structures~\cite{huang2021block} in VVC. Although the partitioning structures become more flexible and efficient, the rectangular partitioning cannot represent arbitrary shape for real objects with a uniform motion vector~(MV) in a rectangular block. In particular, blocks containing object boundaries usually consist of regions moving in distinct directions. Consequently, a single MV cannot capture the real motion characteristics of each region, which may result in high prediction errors. 
	
\begin{figure}[!t]
		\centering
		\includegraphics[width=1.5in]{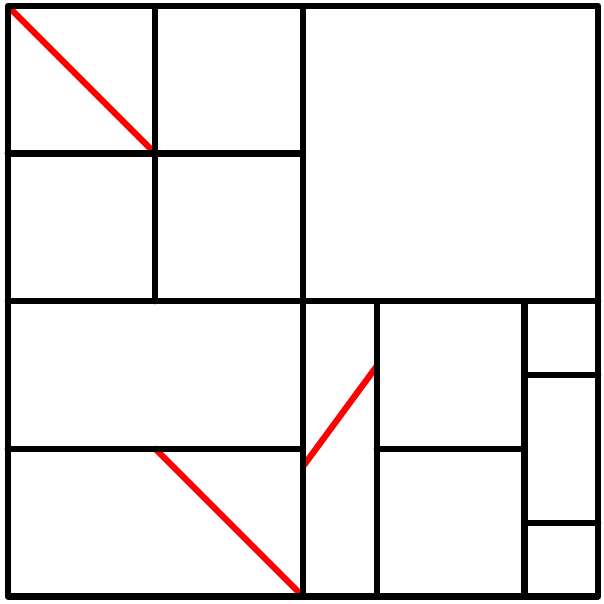}
		\caption{An example of QT-MTT based geometric partitioning scheme. All leaf node CUs of QT-MTT can be further geometrically partitioned by a straight line.}
		\label{partitioning}
		\vspace{-3mm}
	\end{figure}

	To improve the representation capability of block partitioning, geometric partitioning schemes have been intensively studied. During the development of AVC, HEVC and VVC, geometric partitioning was initially applied as a part of the block partitioning structure~\cite{kondo2005motion, karczewicz2010video,chen2010geometry,C301,L0125,L0417,bross2019general}. Taken the geometric partitioning in VVC as an example, the leaf node coding units~(CU) of the newly-adopted QT-MTT structure can be further geometrically partitioned into two subparts by a straight line, as shown in Fig.~\ref{partitioning}. Then, the encoder performs ME on each subpart to find the best-matched reference block, and each of the two resulting subparts are predicted by motion compensation with an associated MV. To reduce the encoder complexity, many optimization approaches~\cite{ferreira2009efficiency,muhit2009fast,D230,guo2010simplified,bordes2011fast,wang2012complexity} were developed. During the standardization of VVC, a triangle partitioning mode~(TPM)~\cite{L0124}, which was for merge mode only, was adopted. For the merge-mode-based TPM, the MVs of the two subparts are implicitly derived from the motion field of previously-coded CUs rather than estimated by ME. To further improve the motion modeling accuracy of block partitioning, a new geometric partitioning~(GEO) scheme~\cite{P0068}\cite{mode140} only for merge mode was proposed. Compared to TPM, GEO employs more fine granularity partitioning patterns. There were up to 140 partitioning modes in the initial version of GEO. Several optimization methods~\cite{mode84, P0107, Q0059} were investigated to improve the software and hardware friendliness. At last, GEO~\cite{Q0806} for merge mode with 64 partitioning modes was adopted by VVC, which replaced the previously-adopted TPM. 
	
\begin{figure}[t!]
		\begin{center}
			\subfigure[]{
				\includegraphics[width=1.037in]{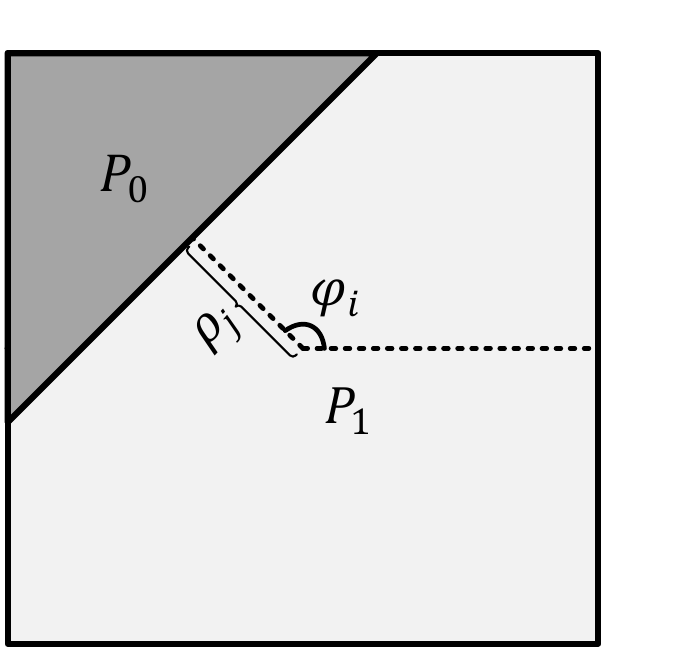}
			}
			\subfigure[]{
				\includegraphics[width=1.037in]{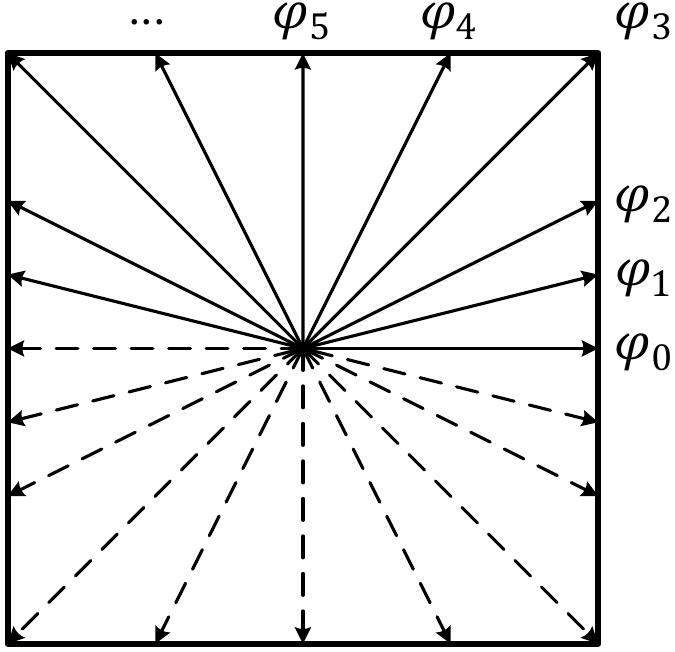}
			}
			\subfigure[]{
				\includegraphics[width=1.037in]{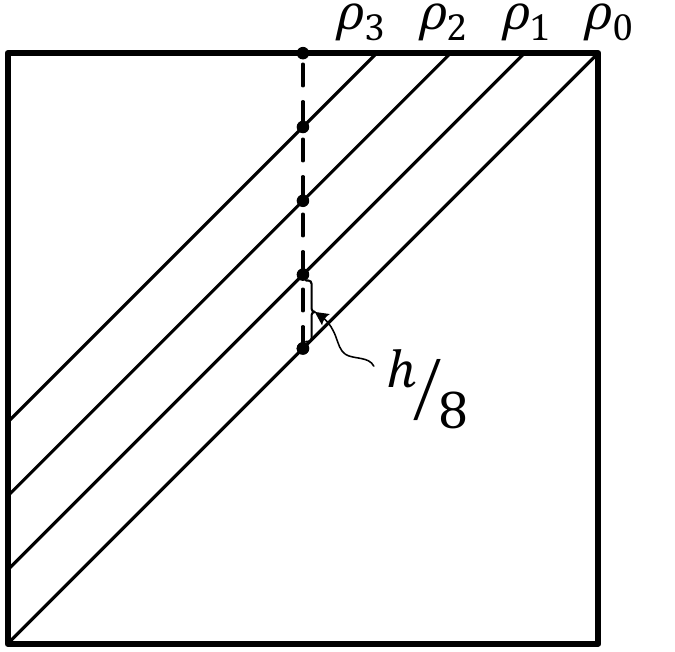}
			}
			\caption{(a)~An example of GEO partitioning mode; (b)~quantized angle $\varphi_{i}$; (c)~quantized offset $\rho_{j}$.}
			\label{distance}
		\end{center}
		\vspace{-3mm}
	\end{figure}

	The GEO adopted in VVC splits a CU into two subparts labeled with $P_0$ and $P_1$ by a straight line, as shown in Fig.~\ref{distance}(a). The straight line is described by the angle $\varphi$ and the distance offset $\rho$~\cite{gao2020geometric}. As shown in Fig.~\ref{distance}(b), the angle $\varphi$ is quantized into 20 discrete angles $\varphi_i$ with the angle of $[0, 2\pi)$ symmetrically divided. The distance offset $\rho$ of each angle is quantized into $\rho_j$ ($j \in \{0, 1, 2, 3\}$) depending on the width $w$ and height $h$ of the CU, as depicted in Fig.~\ref{distance}(c). 
	
	For the merge-mode-based GEO, the partitioning mode decision and the MV selection are vital for the rate-distortion performance. In particular, the partitioning mode is selected for a CU from the pre-defined 64 candidates, and an index~($0\sim63$) is signaled by fixed-length binarization to indicate the selected mode. Regarding the MV, it is selected from the merge candidate list~(MCL). Subsequently, two indices corresponding to the ultimate MVs in the MCL are signaled for $P_0$ and $P_1$ by Truncated Unary Code, respectively. This binarization design is based on the assumption that the candidate at the beginning of the MCL is more likely to be selected as the best one. Generally, the MCL is constructed by including the motion information of previously-coded CUs. The GEO MCL construction process is inherited from the regular MCL, which is initially designed for the conventional rectangular CUs.
	
	Though the newly adopted GEO has higher flexibility to capture the motion field of regions containing moving objects, its indiscriminate coding process for all CUs limits the coding efficiency. First, the number of GEO partitioning modes increases at the expense of increased mode signaling cost, i.e., 6 bits are consumed for each GEO-coded CU. Second, GEO is designed for the CUs containing distinct motion displacement. The preferable MVs for each non-rectangular subpart might be different from each other and also different from the conventional rectangular blocks. Inheriting the MCL construction scheme designed for rectangular CUs limits the adaptability and cannot predict preferable MVs efficiently.
	
	In this paper, we aim to further explore the principle behind GEO towards an adaptive and efficient methodology for side information coding. The main contributions are summarized as follows,
	\begin{itemize}
		\item[1)] We study and theoretically analyze the characteristics of partitioning mode decision and motion field correlation. Through these analyses, we uncover where the coding gain for GEO comes from, i.e., better coding of object boundaries and the corresponding motion field. These characteristics and correlation also motivate us to further improve the GEO coding performance by adaptively designing the signaling approach of side information.
		
		\item[2)] Based on the observations mentioned above, we develop a spatio-temporal correlation guided geometric partitioning scheme~(STGEO) to further improve the coding efficiency of VVC. In particular, we first predict the STGEO partitioning modes by leveraging edge information and history STGEO modes of adjacent blocks, and then adaptively construct the MCL according to off-line trained selection probabilities for merge candidates. Consequently, the partitioning modes and merge candidates with higher selection probabilities can be represented by fewer bits. Statistically speaking, the bits consumed for the side information representation can be reduced.  
		
		\item[3)] Regarding the coding performance, the proposed method achieves significant improvements over GEO with 6\% and 8\% decoding time increment under Random Access~(RA) and Low Delay B~(LDB) configurations in the VVC reference software VTM-8.0~\cite{VTM8.0}. 
	\end{itemize}
	
	The remainder of this paper is organized as follows. Section~\ref{analyses} analyzes the characteristics of partitioning mode decision and motion field correlation, then introduces the framework of the proposed STGEO scheme. Section~\ref{cont1} and Section~\ref{cont2} introduce the main contributions of STGEO, i.e., most probable STGEO mode prediction and probability-based merge candidate list inference. Section~\ref{experimental} shows the experimental results and discussions. Finally, Section~\ref{conclusion} concludes this paper.
	
	\section{Analyses and Motivations} \label{analyses}
	Although the GEO adopted by VVC has higher flexibility in describing the motion field, it causes a heavy burden in signaling the side information, including one GEO mode index and two merge candidate indices. To improve the coding performance of GEO, the key ingredient lies in how to reduce the overhead. In view of this, we investigate the characteristics of partitioning mode decision and motion field correlation in this section.
	
	\begin{figure}[t!]
		\begin{center}
			\noindent
			\subfigure[\textit{BQMall}~(the $82^{nd}$ frame, RA)]{
				\includegraphics[width=3.3in]{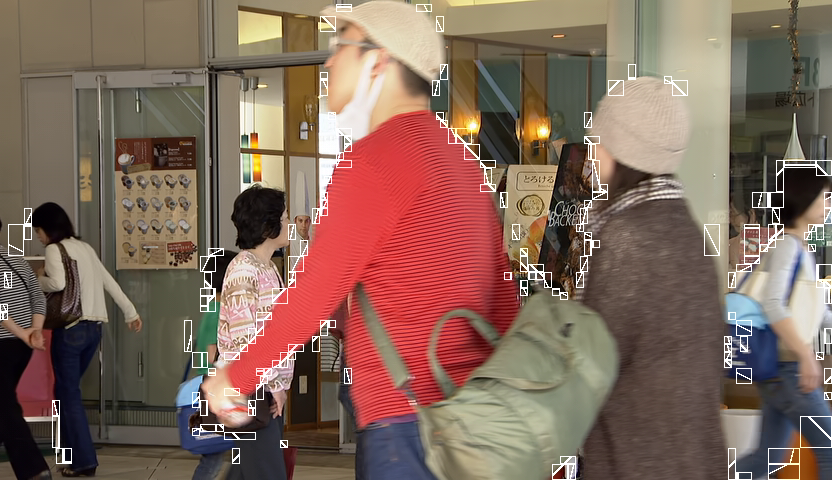}
			}	
		    \subfigure[Motion field of \textit{BQMall}~(the $82^{nd}$ frame, RA)]{
			\includegraphics[width=3.3in]{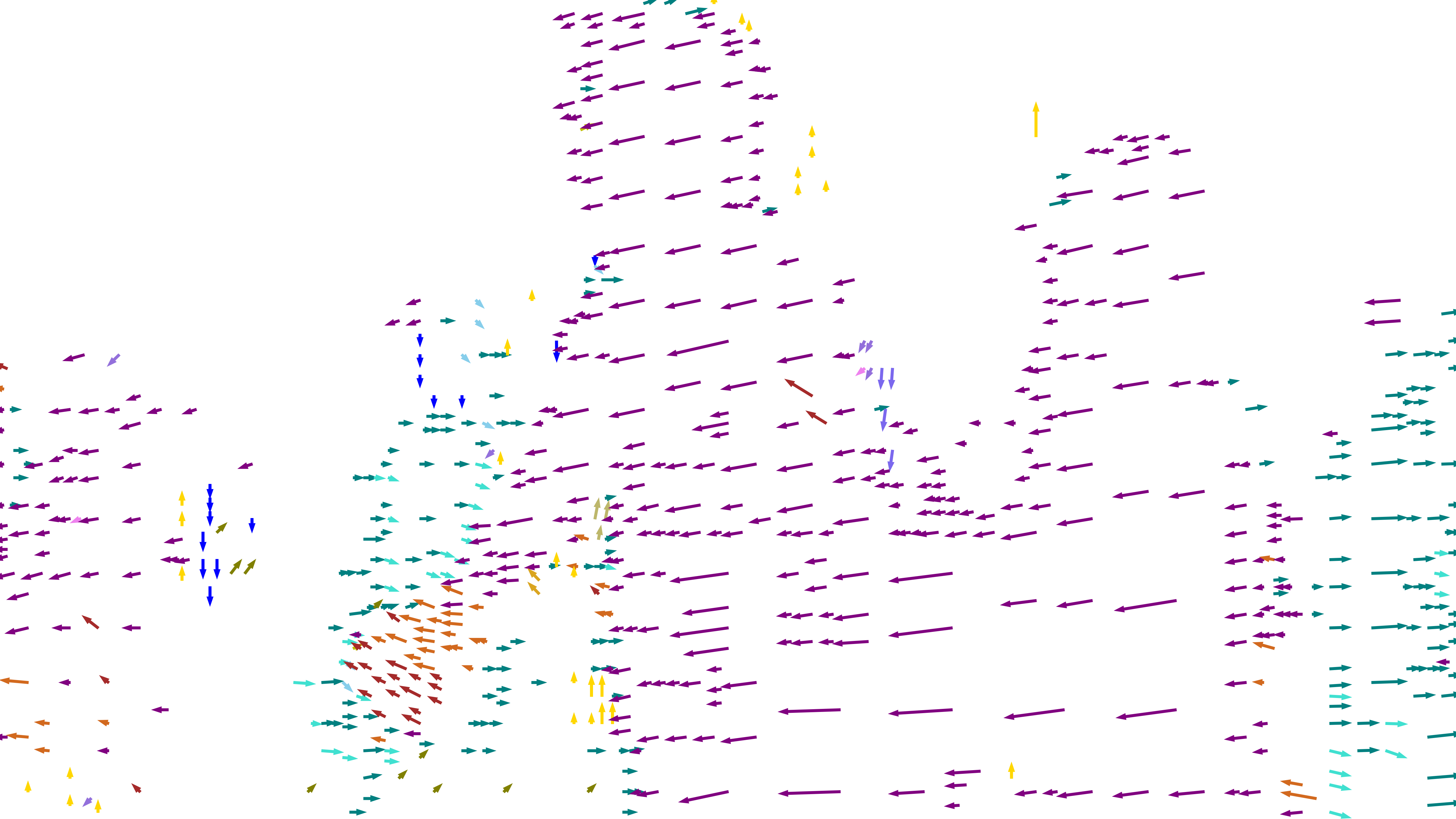}
		     }
			\caption{Illustration of GEO partitioning results and motion field in VTM-8.0. The length of the lines in (b) represents the intensity of motion, and the different colors indicate the motion directions.}
			\label{dec_partition}
		\end{center}
		\vspace{-3mm}
	\end{figure}
	
	To investigate the properties of GEO in describing the motion field, an illustration of the partitioning results and its corresponding motion field is provided in Fig.~\ref{dec_partition}. From Fig.~\ref{dec_partition}(a), one can discern that GEO is frequently adopted around the object boundaries, especially boundaries of moving objects. Given the continuity of object boundaries, the GEO partition line also tends to be piecewise-continuous, such as the boundaries of persons in Fig.~\ref{dec_partition}(a). Furthermore, the motion field of a frame is generally segmented by the boundaries of moving objects as illustrated in Fig.~\ref{dec_partition}(b). The main reason is that objects generally exhibit movement relative to a static background or other moving objects. Consequently, the two subparts in a GEO-coded CU usually have different MVs. The correlation between the MV of each subpart and that of the previously-coded CUs is also distinct for different GEO modes. Hence, the current MCL construction, assuming that all the GEO-coded subparts have the same motion field correlation as rectangular CUs, is not optimal for GEO. 
	
	To further validate the correlation between GEO-coded blocks and object boundaries in a statistically-sound way, we propose to investigate the effectiveness of GEO from the prediction error level. We use $p_r$ to denote the prediction error, which is computed by,
	\begin{equation}\label{resi}
	p_r(x,y) = |p_{rec}(x,y) - p_{pre}(x,y)|,
	\end{equation}
	where $p_{rec}$ and $p_{pre}$ denote the signals of the reconstructed and prediction luma samples, respectively. $(x,y)$ is the location of a sample in a frame. $p_{r1}$ and $p_{r2}$ are used to represent the prediction error of frames coded with and without GEO, respectively. Consequently, $\Delta p_r(x,y)$ can roughly represent the prediction error reduction introduced by GEO of the sample with location $(x,y)$, which is calculated by, 
	\begin{equation}
	     \Delta p_r(x,y) = p_{r2}(x,y) - p_{r1}(x,y).
	\end{equation}
	
	As we mentioned before, GEO is designed to better describe the discontinues motion field. But the motion characteristics are difficult to capture, so we use object boundaries to represent the motion field boundaries. To derive the object boundaries, we conduct the conventional \textit{Canny} detector~\cite{canny1986computational} on the reconstructed luma samples $p_{rec}$. The two thresholds in the \textit{Canny} function are empirically set to be 35 and 70. Then the \textit{dilation} operation is conducted on the canny-detected edge map with the parameter set as 5. With the dilated edge map, pixels in one frame $F$ can be divided into two subsets, edge pixel subset $f_E$ and background pixel subset $f_B$. Consequently, the prediction error reduction $\Delta p_r$ can be classified into two types, $\Delta p_{rE}$ and $\Delta p_{rB}$, corresponding to the edge pixel subset and background pixel subset respectively. $\Delta p_{rE}$ can represent the influence of GEO on edge pixels. We use $\Omega_E$ and $\Omega_B$ to represent the location of edge pixels and background pixels, respectively. $\Delta p_{rE}$ and $\Delta p_{rB}$ are calculated by,
	\begin{equation}\label{edgePixel}
	\Delta p_{rE} = \sum_{(x,y)\in \Omega_{E}}{\Delta p_r(x,y)},
	\end{equation}
	
	\begin{equation}\label{backPixel}
	\Delta p_{rB} = \sum_{(x,y)\in \Omega_{B}}{\Delta p_r(x,y)}.
	\end{equation}
	
	To better illustrate the effectiveness of GEO, we propose to use the average prediction error change per pixel. Assuming $N_E$ and $N_B$ represent the number of pixels in the subset $f_E$ and $f_B$, the average prediction error change per pixel $\hat{\Delta p_{rE}}$ and $\hat{\Delta p_{rB}}$ for edge and background subsets are calculated by,
	
	\begin{figure*}[!t]
		\begin{center}
			\noindent
				\includegraphics[width=7.15in]{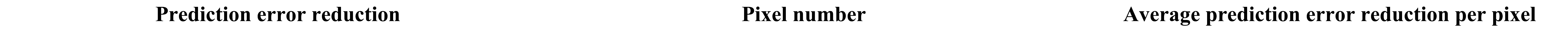}\\
				\includegraphics[width=7.15in]{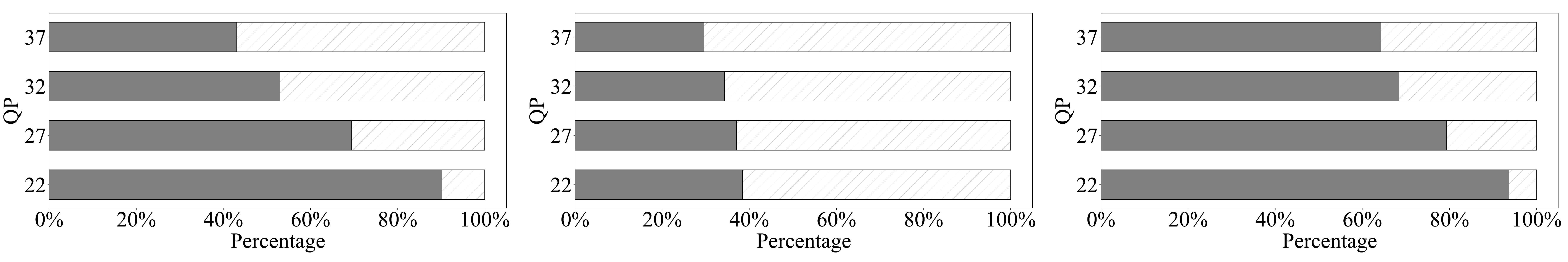}\\
				\bigskip
				\includegraphics[width=7.15in]{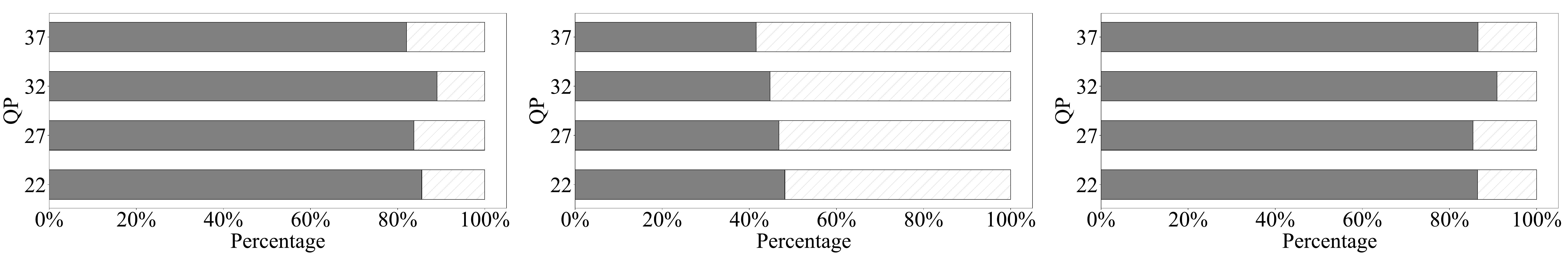}\\
				\includegraphics[width=7.15in]{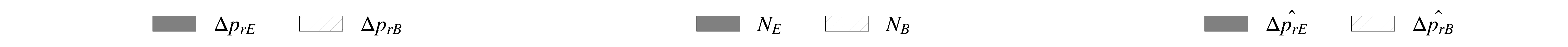}
			\caption{Illustration of the influence of GEO on prediction error. The figures in the first column represent the influence of GEO on the prediction error of edge area $\Delta p_{rE}$ and background area $\Delta p_{rB}$. The second column is the percentage of pixels in edge area $N_{E}$ and the background area $N_{B}$. The figures in the last column are the average prediction error change per pixel of edge area $\hat{\Delta p_{rE}}$ and background area $\hat{\Delta p_{rB}}$, which are normalized for better visualization. Two rows of figures indicate the analysis of \textit{Cactus} and \textit{BQMall}, respectively.}
			\label{edge_back_residual}
		\end{center}
		\vspace{-3mm}
	\end{figure*}
	
	\begin{figure}[!t]
		\begin{center}
			\noindent
			\subfigure[]{
				\includegraphics[width=3.15in]{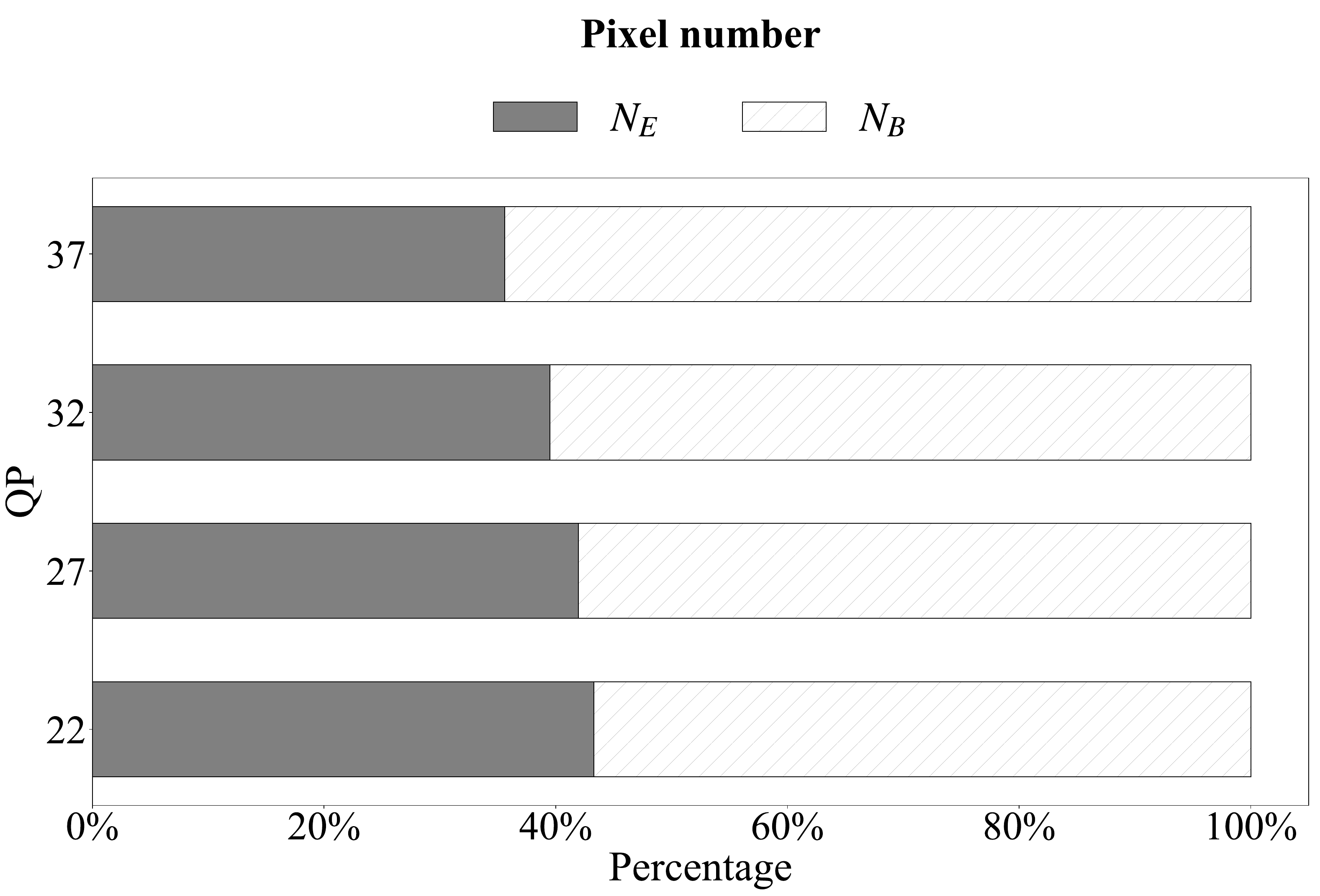}
			}
			\subfigure[]{
				\includegraphics[width=3.15in]{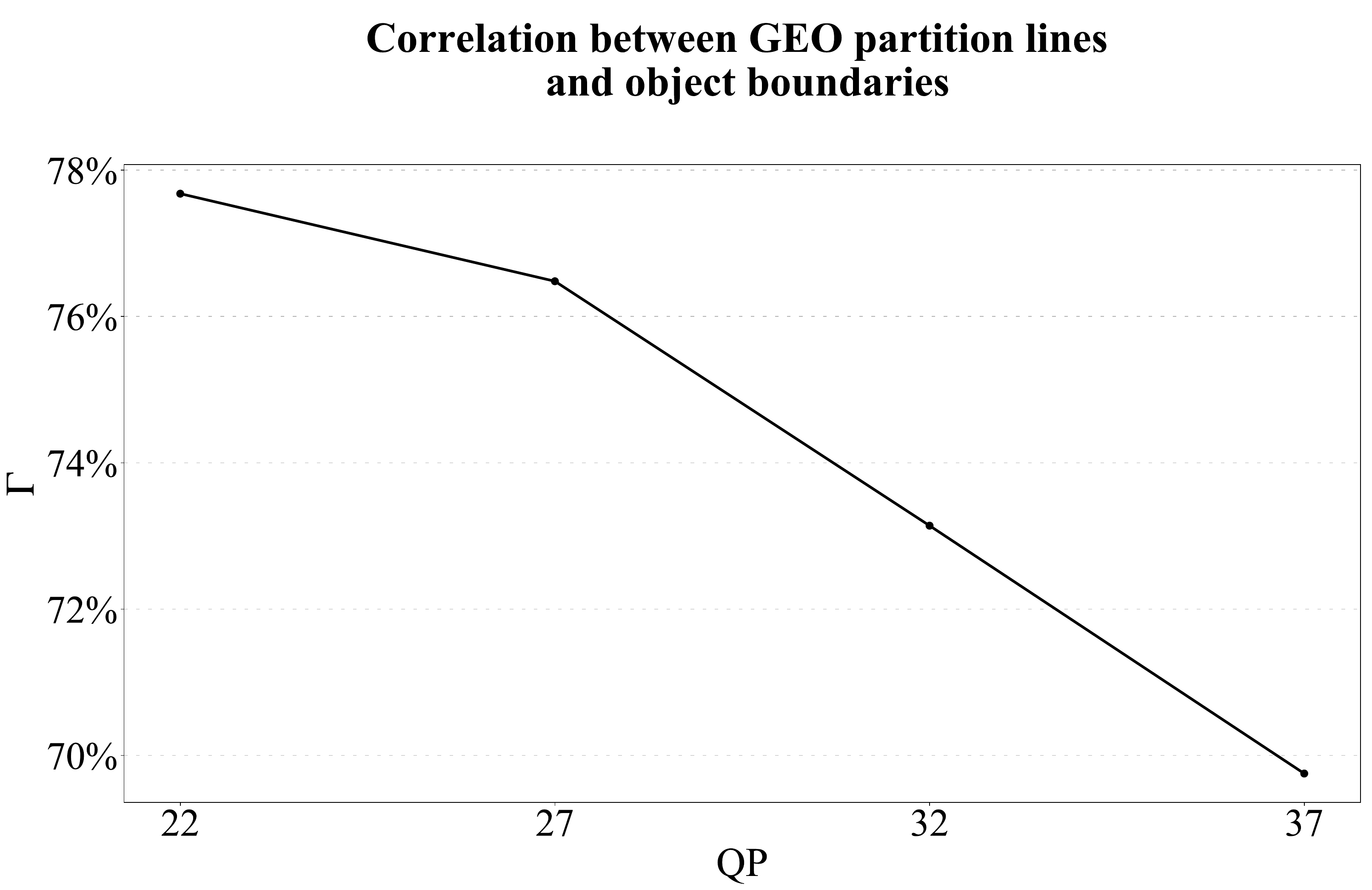}
			}
			\caption{Illustration of correlation $\Gamma$ between GEO partition lines and object boundaries. (a)~the percentage of pixels in edge area $N_{E}$ and background area $N_{B}$; (b)~the percentage of pixels on GEO partition lines that belong to the edge area.}
			\label{edge_back_ratio}
		\end{center}
		\vspace{-3mm}
	\end{figure}

	\begin{equation}\label{per-pixel}
	\hat{\Delta p_{rE}} = \frac{\Delta p_{rE}}{N_E},
	\end{equation}
	
	\begin{equation}\label{per-pixel2}
	\hat{\Delta p_{rB}} = \frac{\Delta p_{rB}}{N_B}.	
	\end{equation}
	\noindent For better visualization, $\hat{\Delta p_{rE}}$ and $\hat{\Delta p_{rB}}$ are normalized.
	
	To collect the statistics mentioned above, we conduct an experiment in VTM-8.0 using two sequences~(\textit{Cactus} and \textit{BQMall}) under LDB configuration. The quantization parameters~(QP) are set as \{22, 27, 32, 37\}, and the duration of each test sequence is 10 sec. As shown in Fig.~\ref{edge_back_residual}, the results exhibit that the prediction error reduction caused by GEO is mainly because of the better prediction of object boundaries. In particular, on average 77\% prediction error reduction of GEO belongs to the edge area. However, the edge pixels account for no more than 40\% of the whole test sequences. The normalized average prediction error change per pixel in the third column of Fig.~\ref{edge_back_residual} can better illustrate this observation. So we can conclude that the coding performance introduced by GEO is mainly due to the improvement on prediction accuracy of object boundaries. However, the bits cost for representing the quantized geometric partition line is a non-negligible burden, i.e., 6 bits for each CU. This significantly limits the overall coding performance. Based on the assumed correlation between GEO partition lines and object boundaries in Fig.~\ref{dec_partition}, the most probable GEO modes for a particular CU are predictable, which can guide the mode signaling design. To validate this correlation, we propose to use $\Gamma$, i.e., the percentage of pixels on GEO partition lines belonging to the edge subset $f_E$. Let $\Omega_L$ represent the location of pixels on GEO partition lines. The percentage $\Gamma$ can be calculated by,
	\begin{equation}
	\Gamma = \frac{Card(\Omega_L \cap \Omega_{E})}{Card(\Omega_L)},
	\end{equation}
	\noindent where $Card(\cdot)$ is the number of elements in the set. The average results of the two sequences~(\textit{Cactus} and \textit{BQMall}) are shown in Fig.~\ref{edge_back_ratio}. It is obvious that most of the pixels on the GEO partition lines are the edge pixels, which indicates that partition lines have a high correlation with object boundaries. 

	\begin{figure*}[t!]
		\begin{center}
			\noindent
			\includegraphics[width=5.0in]{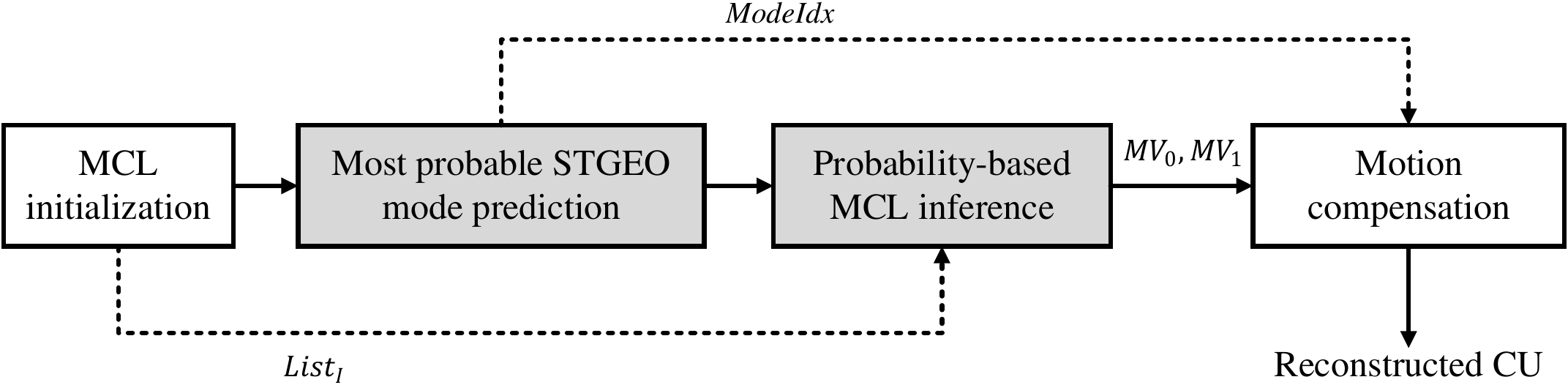}
			\caption{Illustration of the framework of the proposed STGEO scheme at the decoder, with the dotted arrows representing the coding information and the gray boxes highlighting our contributions.}
			\label{vvc-frame}
		\end{center}
		\vspace{-3mm}
	\end{figure*}

	Based on the observations mentioned above, we develop a STGEO scheme to further improve the coding efficiency of VVC. The proposed STGEO is applied as a merge mode in the inter prediction on the CU level. The framework of the proposed STGEO is illustrated in Fig.~\ref{vvc-frame}. It includes \textit{MCL initialization}, \textit{most probable STGEO mode prediction}, \textit{probability-based MCL inference} and \textit{motion compensation}. First, given an STGEO-coded CU, the MCL is initialized by reusing the regular MCL construction process. Then, with the $List_{I}$, the STGEO mode set is split into two subsets, high-probability subset~(HPS) and low-probability subset~(LPS), based on the most probable STGEO mode prediction scheme. Combining the two subsets with the decoded syntax elements in Section~\ref{cont2}, the \textit{ModeIdx} can be derived. The \textit{ModeIdx} indicates the partitioning mode index in the STGEO mode set. Subsequently, with the \textit{ModeIdx} and $List_{I}$, the MCLs of the two STGEO generated subparts are adaptively inferred based on the off-line trained merge candidate selection probability. With the decoded merge candidate indices in Section~\ref{cont2}, $MV_0$ and $MV_1$ are derived for the two subparts, respectively. Finally, motion compensation is performed. 

	The \textit{MCL initialization} is to derive the $List_{I}$ for the \textit{most probable STGEO mode prediction} and \textit{probability-based MCL inference}. The MCL initialization includes two steps, regular MCL construction inherited from the conventional rectangular CUs and uni-directional MCL derivation. The regular MCL $List_R$ is constructed by including the following five types of Motion Vector Predictor~(MVP) in order~\cite{VTM8.0}, i.e., Spatial MVP, Temporal MVP~(TMVP), History-based MVP~(HMVP)~\cite{HMVP}\cite{zhang2019history}, Pairwise average MVP~(PMVP)~\cite{pairwise} and Zero MVP. For Spatial MVP, a maximum of four merge candidates are selected among candidates located in the positions ``up'', ``left'', ``up-right'', ``down-left'' and ``up-left.'' This inheritance method can enable the decoder to reuse the MCL construction module of rectangular CUs for STGEO, which can simplify the decoder design. Then uni-directional MCL derivation is conducted to convert bi-directional candidates in the $List_{R}$ to uni-directional ones for the sake of worst-case memory bandwidth reduction~\cite{gao2020geometric}. 
	
	\textit{Most probable STGEO mode prediction} and \textit{probability-based MCL inference} are the main contributions of this paper. They are detailed in Section~\ref{cont1} and Section~\ref{cont2}, respectively.
	
	\section{Most Probable STGEO Mode Prediction} \label{cont1}
	Based on our observation and analysis in Section~\ref{analyses}, blocks containing moving object boundaries with discontinuous motion fields are more likely to be coded by GEO, and the GEO partition lines have a high correlation with the boundaries. Therefore, with the boundary location, the STGEO modes with high selection probabilities are predictable. Based on the predicted STGEO modes, STGEO mode set can be split into two subsets, HPS containing the partitioning modes with high preference and LPS consisting of the remaining low-preference modes. The STGEO modes in the HPS can be represented with smaller indices, and fewer bits are consumed, leading to better compression performance. Consequently, the critical ingredient lies in how to predict the high-preference STGEO modes~\cite{myICME}. There are some previous works~\cite{blaser2017geometry,J0023,blaser2018geometry} aiming at predicting the GEO modes, in which the GEO partition line of the current block is predicted based on that of the block in reference frames. Different from that, our method utilizes the boundary information of the reference blocks. The main reason is that the reference block may be not exactly partitioned by geometric partitioning mode. Boundary information can overcome this shortcoming and the memory requirements for storing the partitioning modes can be avoided. In our method, as the current CU is unavailable at the decoder, \textit{reference block derivation} process is conducted first.
	Then \textit{edge detection} is applied on the reference block to locate the object boundaries. According to the detected edge information and STGEO modes of neighboring CUs, \textit{HPS construction} is performed by including the predicted high-preference STGEO modes. Details are given below.
	
	\begin{figure}[t!]
		\begin{center}
			\noindent
			\includegraphics[width=3.35in]{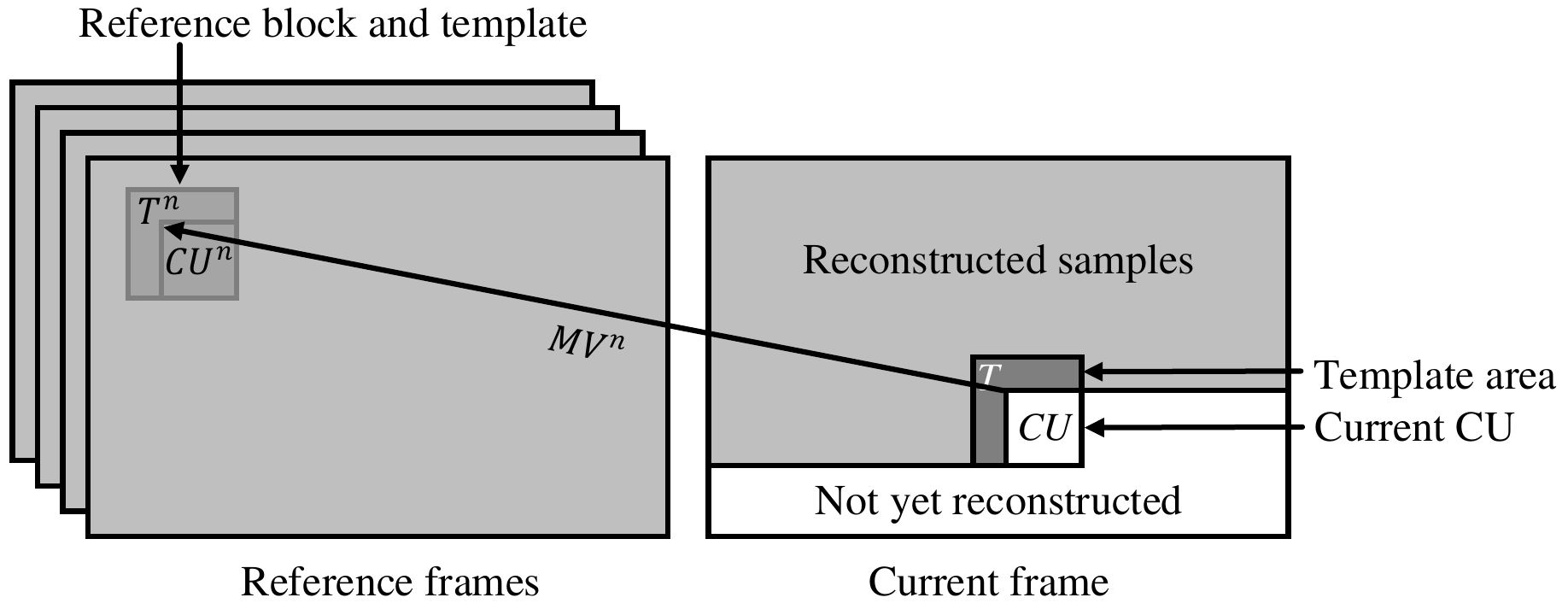}
			\caption{Illustration of the \textit{reference block derivation} process.}
			\label{templateMatching}
		\end{center}
		\vspace{-3mm}
	\end{figure}
	
	\subsection{Reference Block Derivation}
	Due to the temporal correlations of moving objects, the object boundary information can be obtained from the reference frames~\cite{wang2019three}. Herein, we derive the reference block by a template matching approach, which has been intensively studied in intra prediction~\cite{tan2006intra} and inter prediction~\cite{sugimoto2004inter}. As shown in Fig.~\ref{templateMatching}, we first derive the reference blocks of the current CU by motion compensation using each MV in the $List_{I}$. Then the mean square error~(MSE) between the template area of the current CU~($T$) and that of its reference block~($T^{n}$) is calculated. The MSE is used as the similarity evaluation criteria. Finally, the block with the smallest MSE is selected as the reference block used for the following \textit{edge detection}.
	
	\subsection{Edge Detection}
	To locate the object boundaries in the reference block derived above, a novel low-complexity edge detector is proposed. The main process is similar to the \textit{Canny} detector. First, the Sobel operator is conducted on the reference block to derive the amplitude and angle of each sample. With the amplitude map and angle map, non-maximum suppression is applied to find the locations with the sharpest change of intensity value. Finally, a novel double threshold process is performed to get the final edge map and angle map. For a block $\Omega$ with width $w$ and height $h$, we use the one-dimensional vector $E$ as the edge map after the non-maximum suppression process. $E(\textbf s) (\textbf s\in \Omega)$ represents the amplitude located at position $(s_x, s_y)$ in the edge map. In the double threshold process, the edge map $\hat{E}$ is derived by,

	\begin{figure}[!t]
	\begin{center}
		\noindent
		\subfigure[]{
			\includegraphics[width=1.6in]{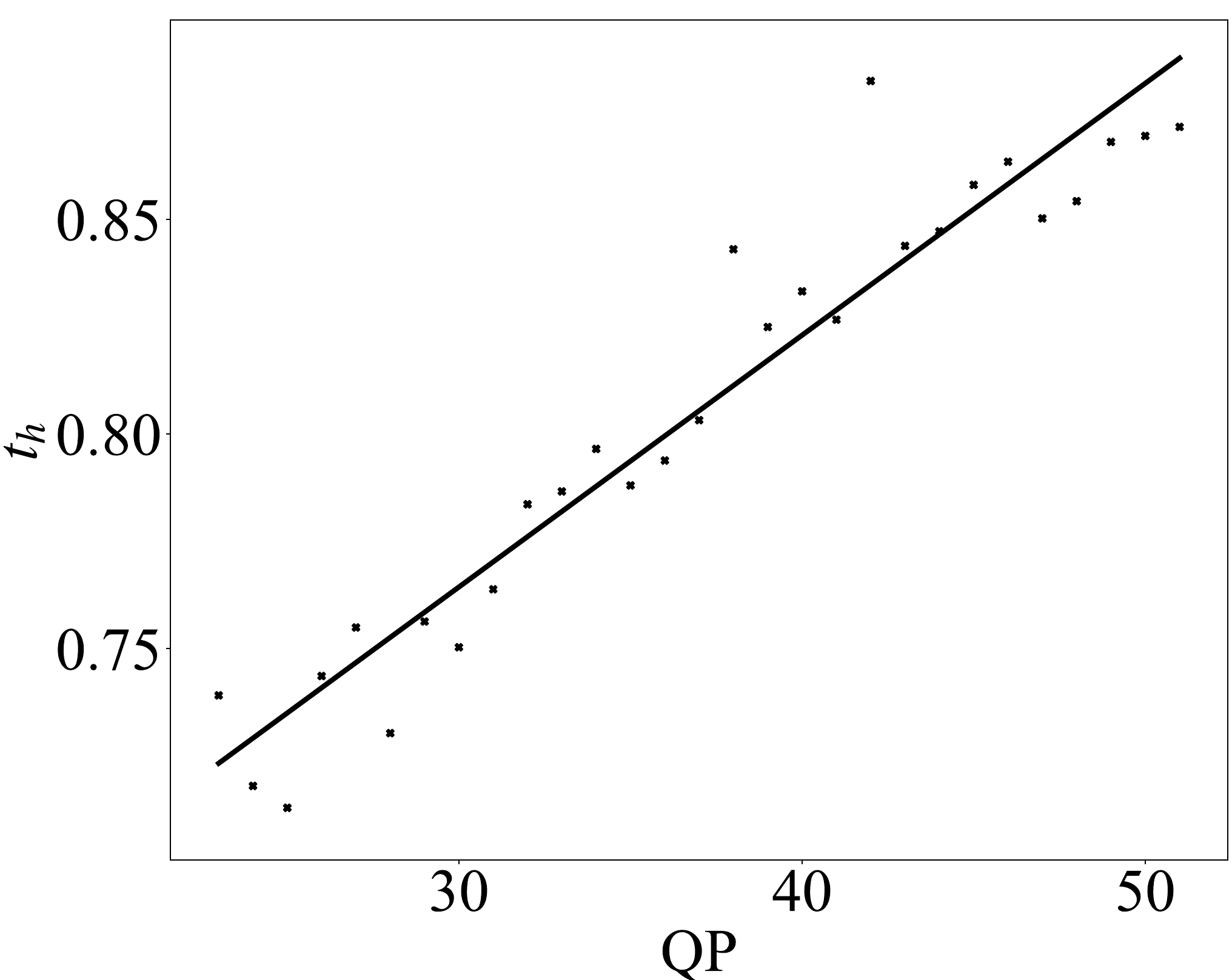}
		}
		\subfigure[]{
			\includegraphics[width=1.6in]{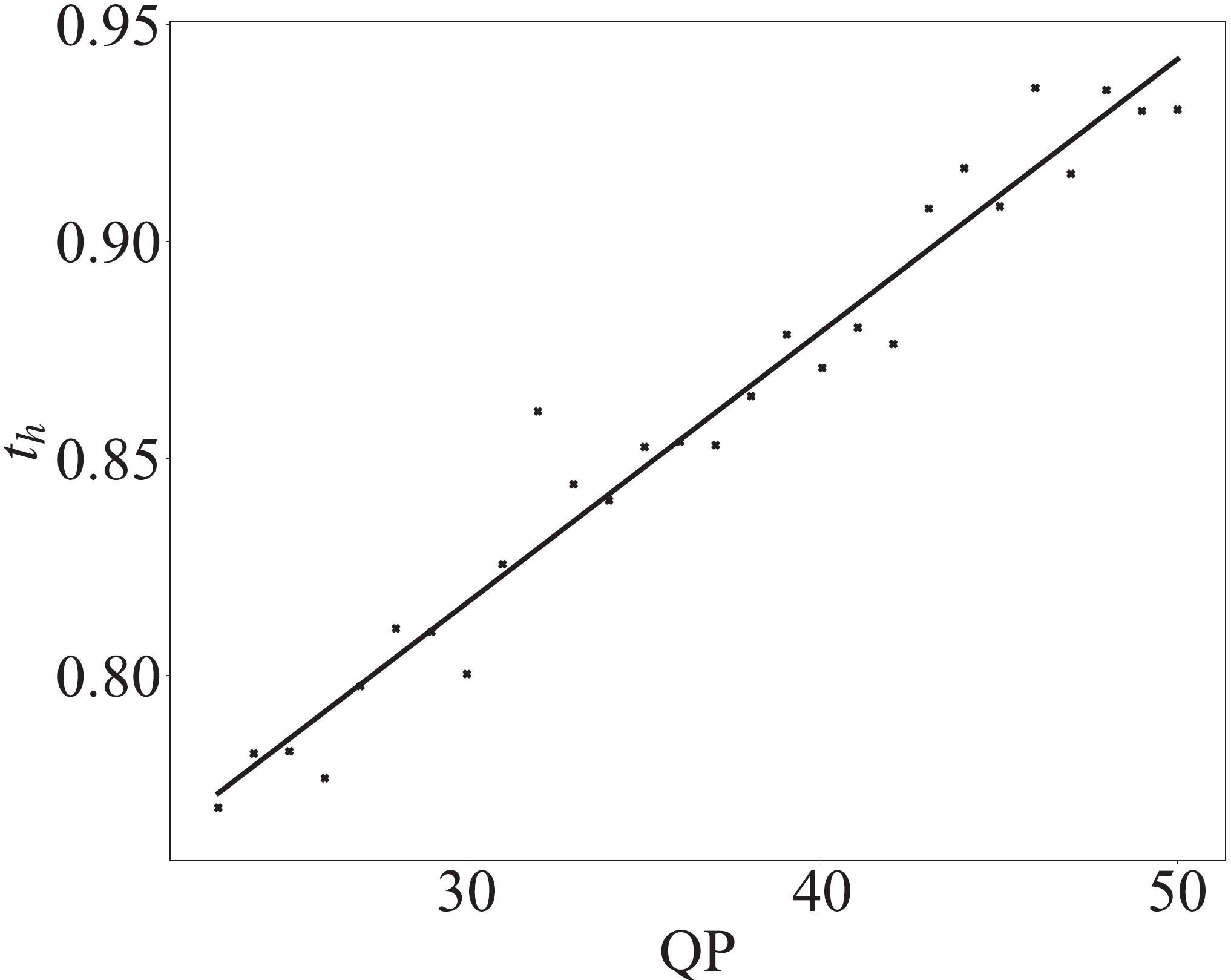}
		}
		\caption{Illustration of the relationship between parameter $t_{h}$ and QP in \textit{edge detection} process. (a)~RA; (b)~LDB.}
		\label{training}
	\end{center}
	\vspace{-3mm}
	\end{figure}

	\begin{equation}
	\hat{E}(\textbf s) = 
	\begin{cases}
	0 & \mathrm{if} \text{ } \mathrm{Condition1} \text{ } \mathrm{or} \text{ } \mathrm{Condition2,}\\
	E(\textbf s) & \mathrm{otherwise.}
	\end{cases}
	\end{equation}
	
	\begin{equation}
	\mathrm{Condition1}: E(\textbf s) < T_{\mathrm{low}},
	\end{equation}
	
	\begin{equation}
	\begin{aligned}
	\mathrm{Condition2}: & E(\textbf s) < T_{\mathrm{high}} &\text{ } \mathrm{and }\\
	& E(\textbf s) \geq T_{\mathrm{low}} &\text{ } \mathrm{and }\\
	& E(\textbf s ^{'}) < T_{\mathrm{low}},
	\end{aligned}
	\end{equation}where $E(\textbf s ^{'})$ represents the amplitude located at $(s_x^{'}, s_y^{'})$. $(s_x^{'}, s_y^{'})$ is the location at the pre-defined neighboring area $\Omega^{'}$ of the current location $(s_x, s_y)$. So the two thresholds $T_{\mathrm{high}}$ and $T_{\mathrm{low}}$ are the core elements for the final edge map. In this work, $T_{\mathrm{high}}$ is calculated by, 
	\begin{equation}\label{edge_map_reorder}
	{E}^{'} = \mathrm{Rank}(E) = \{e_1, e_2, ..., e_{N}\} ,
	\end{equation}
	\begin{equation}\label{thre_high}
	T_{\mathrm{high}} = e_{\lambda}, \mathrm{wherein} \text{ } \lambda = N \times t_{h},
	\end{equation}where $\mathrm{Rank}$ is a sorting process in ascending order. $\{e_1, e_2, ..., e_{N}\}$ represents amplitudes after sorting process. $N$ is the number of pixels in the block $\Omega$. $t_{h}$ is a parameter between 0 and 1 trained off-line. The threshold $T_{\mathrm{low}}$ is calculated by, 
	
	\begin{equation}\label{thre_low}
	T_{\mathrm{low}} = \frac{T_{\mathrm{high}}}{t_{l}},
	\end{equation}where $t_{l}$ is the parameter trained off-line.
	
	\begin{figure*}[ht!]
		\begin{center}
			\includegraphics[width=7.0in]{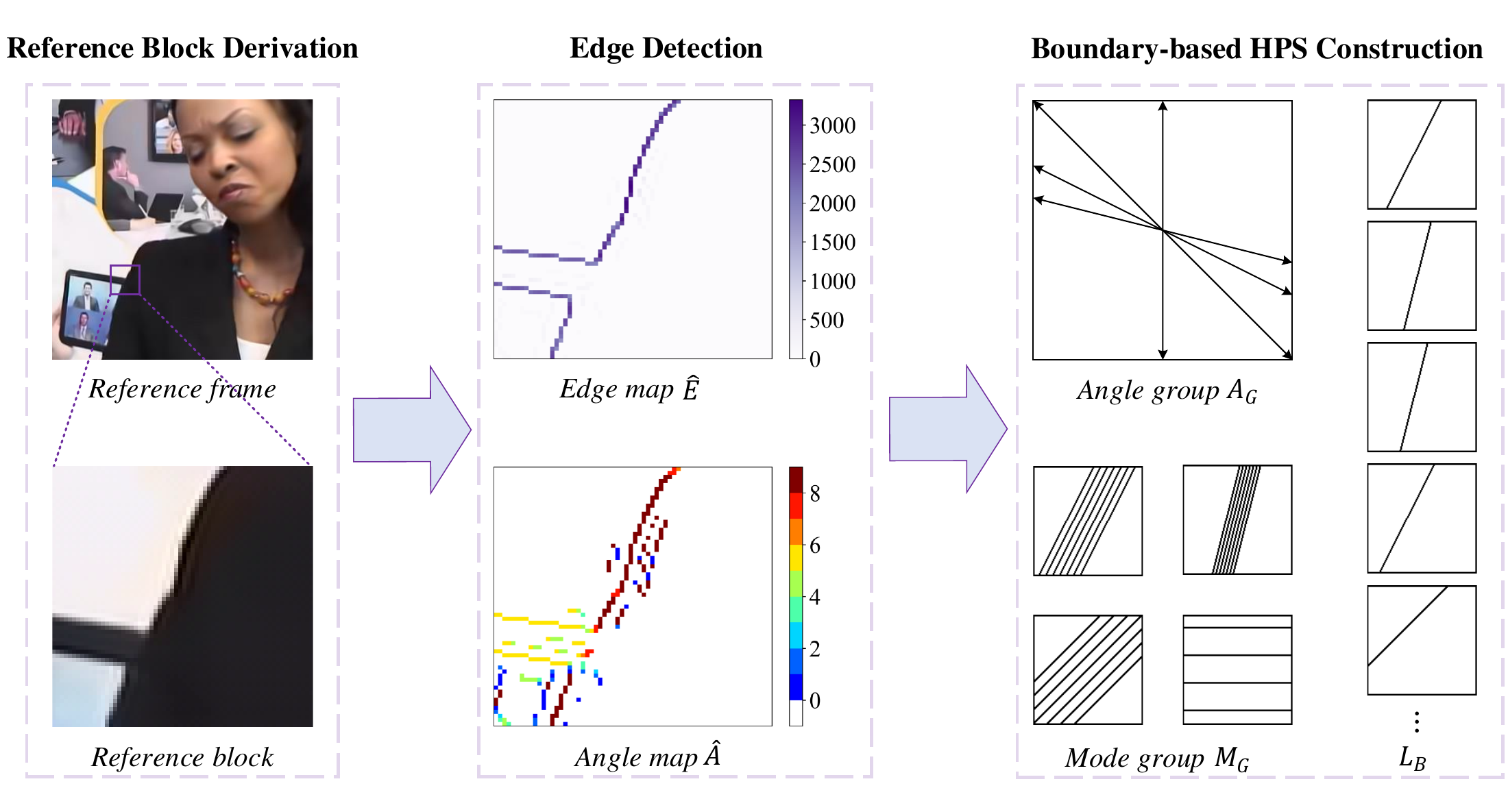}
			\caption{Illustration of the \textit{boundary-based HPS construction} process. Edge map $\hat{E}$ corresponding to the Sobel gradient amplitude is derived in the \textit{edge detection}. Angle map $\hat{A}$ is a quantized map corresponding to the Sobel gradient direction. The most probable angles calculated using $\hat{E}$ and $\hat{A}$ are added in the angle group $A_G$. The mode group $M_G$ consists of STGEO modes corresponding to the $A_G$. $L_B$ is constructed using modes with largest selection probabilities in $M_G$. }
			\label{MPM-list-framework}
		\end{center}
		\vspace{-3mm}
	\end{figure*}
	
	The parameter training process is conducted using the videos and configurations defined in Table \ref{Test_condition2}. QPs are set as \{22, 23, 24, 25, ..., 41, 42\}. The number of coded frames is 16, corresponding to the number of pictures in the Group Of Pictures~(GOP). For each B-frame, we experimentally select the optimal $t_{h}$ corresponding to the highest PSNR of Y component. The results are plotted in Fig.~\ref{training}. Because of the QP Offsets for different B-frames, there may be several QPs for each QP setting. Hence, the number of points in Fig.~\ref{training} is more than 21. It is clear that there is a positive relation between QP and the corresponding best $t_{h}$. So we fit a linear model to represent their correlation. The fitted lines of RA and LDB can be formulated as,
	\begin{equation}
	t_{h} = 0.00586 \times QP + 0.5884\label{threHigh},
	\end{equation}
	\begin{equation}
	t_{h} = 0.00625 \times QP + 0.6288\label{threHigh2}.
	\end{equation}
	
	The parameter $t_{l}$ is trained with $t_{h}$ fixed by \eqref{threHigh} or \eqref{threHigh2}. As there is no obvious rule for this parameter, we set it to be 3.0.
	
	\begin{table}[t!]
		\centering
		\begin{center}
			\caption{Illustration of the sequences and configurations for the proposed STGEO.} \label{Test_condition2}
			\label{condition}
			\begin{tabular}{c c c c}
				\thickhline
				\hline
				{\textbf{Sequence}}     &{\textbf{Resolution}}&{\textbf{FrameCount}}&{\textbf{Configurations}}\\
				\hline
				\textit{TrafficFlow}    & 3840x2160  & 300 & RA\\
				\textit{pku\_girls}     & 3840x2160  & 150 & RA\\
				\textit{Kimono}         & 1920x1080  & 240 & RA, LDB\\
				\textit{ParkScene}      & 1920x1080  & 240 & RA, LDB\\
				\textit{City}           &   832x480  & 600 & RA, LDB\\
				\textit{Crew}           &   832x480  & 600 & RA, LDB\\
				\textit{vidyo1}         &   832x480  & 600 & RA, LDB\\
				\textit{vidyo3}         &   832x480  & 600 & RA, LDB\\
				\textit{BasketballPass} &   416x240  & 500 & RA, LDB\\
				\textit{BQSquare}       &   416x240  & 600 & RA, LDB\\
				\textit{BlowingBubbles} &   416x240  & 500 & RA, LDB\\
				\textit{RaceHorses}     &   416x240  & 300 & RA, LDB\\
				\thickhline
			\end{tabular}
		\end{center}
		\vspace{-3mm}
	\end{table}

	\subsection{High-probability Subset Construction}
	After the \textit{reference block derivation} and \textit{edge detection}, the HPS is constructed. Generally speaking, the HPS construction is an inference problem based on the available information of the previously decoded frames and reconstructed area of the current frame to infer several probable STGEO modes. According to our analyzed correlations between object boundaries and partition lines, the boundary-based high-preference STGEO modes are included first. Then history-based high-preference STGEO modes are added inspired by the observed object boundary continuity. If the HPS is not filled, default modes that show relative high selection probabilities are added. Consequently, three construction subprocesses are needed, including the \textit{boundary-based HPS construction}, \textit{history-based HPS construction}, and \textit{default HPS construction}. The maximum number of elements in the HPS is 15 in this paper. Duplication checking is conducted when constructing the HPS, such that only unique modes are included in it.
		
	The first step, \textit{boundary-based HPS construction}, is to derive high-preference STGEO modes based on the assumption that the STGEO modes with higher boundary coincidence degree are more likely to be selected as the best one. The boundary coincidence degree of a STGEO mode is represented by the average amplitude along its corresponding partition line. Details are shown in Fig.~\ref{MPM-list-framework}. For a block $\Omega$ with width $w$ and height $h$, let $\hat{E}(\textbf{s})$ and $A(\textbf{s})$ denote the edge and angle located at $(s_x, s_y)$ in the edge map $\hat{E}$ and angle map $A$. As STGEO mode is described by the quantized angle $\varphi_{i}$ and distance $\rho_{j}$, the angles in the angle map are firstly converted to the index of their nearest angle in the pre-defined angle candidates $\{\varphi_{0}, \varphi_{1}, \varphi_{2}, ..., \varphi_{19}\}$. The converted angle map is represented by $\hat{A}$. Then, to derive the most probable angles, each amplitude $\hat{E}(\textbf{s})$ in the edge map is classified into 20 categories according to its corresponding angle $\hat{A}(\textbf{s})$. We use $\Omega_i$ to represent the location of amplitudes corresponding to the angle $\varphi_i$. Then the selection probability $SP_i$ of the angle $\varphi_{i}$ is derived by,
	
	\begin{figure}[!t]
		\centering
		\includegraphics[width=1.5in]{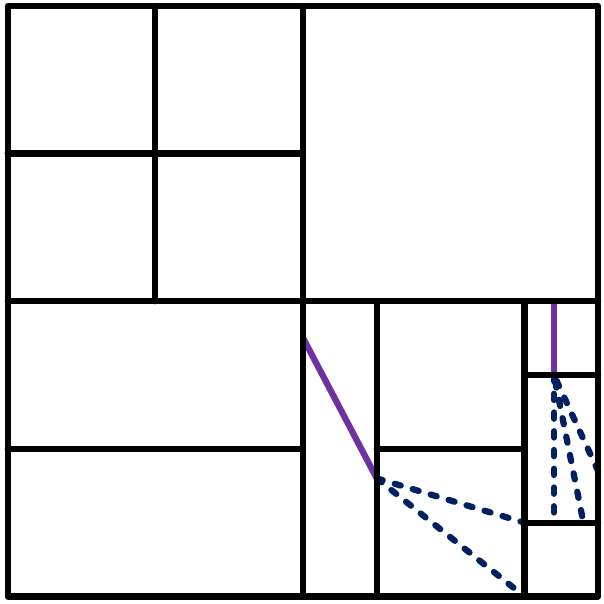}
		\caption{An example of history-based HPS construction. Purple solid lines are STGEO partition lines, and blue dash lines represent the probable modes predicted based on the neighboring STGEO mode.}
		\label{splite_connect}
		\vspace{-3mm}
	\end{figure}
	
	\begin{equation}\label{angProbability}
	SP_i = \sum_{\textbf{s}\in \Omega_i}{\hat{E}(\textbf{s})}.
	\end{equation}
	The two angles with the largest selection probabilities are added to the angle group $A_G$, and the neighboring angles of the largest-probability angle are also included in $A_G$. The main reason for this design is that there may be several boundaries in the reference block, and the primary angle with the largest selection probability may be not precise enough. Subsequently, the STGEO modes corresponding to $A_G$ are added in the mode group $M_G$. Then the boundary coincidence degree $D_k$ of each mode $k$ in $M_G$ is expressed by the average amplitudes along the partition line area $\Omega_l$, 
	
	\begin{equation}\label{modeProbability}
	D_k = \frac{\sum_{\textbf{s}\in \Omega_l}\hat{E}(\textbf{s})}{N},
	\end{equation}
	where $N$ is the number of pixels in $\Omega_l$. Similar to the construction law of angle group $A_G$, the $M$ modes with the largest $D_k$ are added in the boundary-based HPS $L_B$ in order. The maximum modes added in the $L_B$ is 12.
	
	The second step, \textit{history-based HPS construction}, is to construct HPS based on the high spatial correlation due to the continuity of the object boundaries, as illustrated in Fig.~\ref{dec_partition}. Specifically, the history-based high-preference STGEO modes are derived based on the STGEO partitioning modes of the neighboring CUs. Different from \cite{blaser2017geometry,J0023,blaser2018geometry} in which the partition line is predicted by simply linearly extending the partition line of a neighboring block into the current block, we enable more intersected lines. The main reason is that object boundaries of natural videos rarely follow a purely straight line. As shown in Fig.~\ref{splite_connect}, blue dash lines represent the probable modes predicted based on the neighboring STGEO-coded blocks split by purple solid lines. The predicted partition lines and the neighboring partition lines intersect on the side where the two blocks overlap. In particular, the STGEO modes whose corresponding partition lines intersect with neighboring partition lines are added in the history-based HPS. If none of the neighboring CUs is coded by STGEO, the history-based HPS is null.
	
	\begin{figure}[t!]
		\begin{center}
			\noindent
			\includegraphics[width=3.2in]{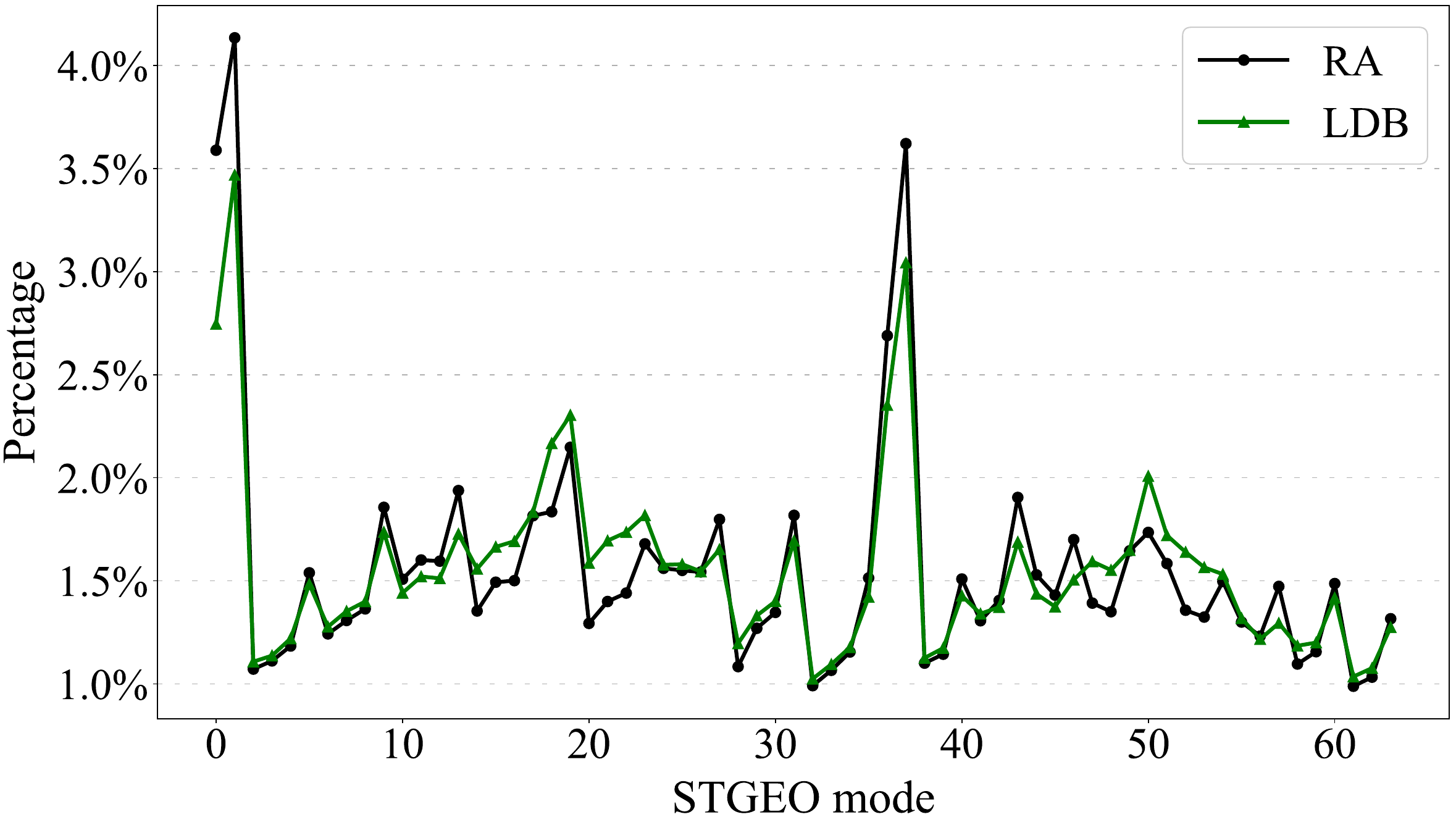}
			\caption{Illustration of the percentage of STGEO index.}
			\label{default_list}
		\end{center}
		\vspace{-3mm}
	\end{figure}
	
	The third step, \textit{default HPS construction}, is conducted when the existing high-preference STGEO modes do not fulfill the HPS. To derive the most probable default candidates, we analyze the percentage of STGEO modes using testing conditions in Table \ref{Test_condition2}. For each test sequence, all frames are used for the experiment. QPs are set as \{22, 23, 24, 25, ..., 41, 42\}. The results are shown in Fig.~\ref{default_list}, which exhibit that the percentage of STGEO modes is not equal. In particular, some modes, such as mode 0, 1, 37, and 38, have a much higher percentage to be selected, and their distributions under RA and LDB configurations are slightly different. So we construct the default list using the most selected STGEO modes for RA and LDB, respectively.
	
	\section{Probability-based Merge Candidate List Inference} \label{cont2}
	Due to the Truncated Unary Code binarization approach designed for the merge index, the overhead caused by the merge index signaling largely depends on the order of merge candidates in the MCL. However, based on our observations, the correlation between motion information of the current CU and that of the previously-coded blocks is different between rectangular and STGEO-coded CUs. Hence, inheriting the MCL $List_{I}$ directly is not the optimal option for STGEO. Here, we propose to adaptively infer the MCL based on the off-line trained candidate selection probability for the two subparts, i.e., $P_0$ and $P_1$, respectively. In particular, the merge candidates with higher selection probabilities are put at the front location of the MCL, vise versa. As such, the bits consumed for representing the merge index can be economized.
	
	To measure the merge candidate preference in a quantitative manner, we propose an evaluation criterion, i.e., the signal-level difference in terms of MSE, which is calculated by,
	\begin{equation}
	MSE =  \frac{1}{N} \sum_{(x,y)\in \Omega}{{\big(p_{org}(x,y)-p_{pre}(x,y)\big)}^2}, 
	\end{equation}
	where $\Omega$ is the current CU with width $w$ and height $h$. $N$~($N=w\times h$) is the number of pixels in the current CU. $(x,y)$ is the location of sample in a frame. $p_{org}$ represents the signals of the original luma samples. $p_{pre}$ denotes the luma prediction samples generated by motion compensation using MVs in the MCL. The MV corresponding to smaller MSE means that the correlation between its prediction block and the original block is higher. Hence, its selection probability is larger. Herein, we first verify the efficiency of the proposed evaluation criterion, i.e., MSE. Then the merge candidate preferences of rectangular CUs and STGEO-coded subparts are statistically analyzed in the sense of MSE. Inspired by the merge candidate preferences, the probability-based MCL inference scheme is detailed. All the statistics are collected with the experimental conditions set as Table \ref{Test_condition2}. For each sequence, all frames are used for the experiment. QPs are set as \{22, 23, 24, 25, ..., 41, 42\}.
	
	\begin{figure}[!t]
		\begin{center}
			\noindent
			\includegraphics[width=3.2in]{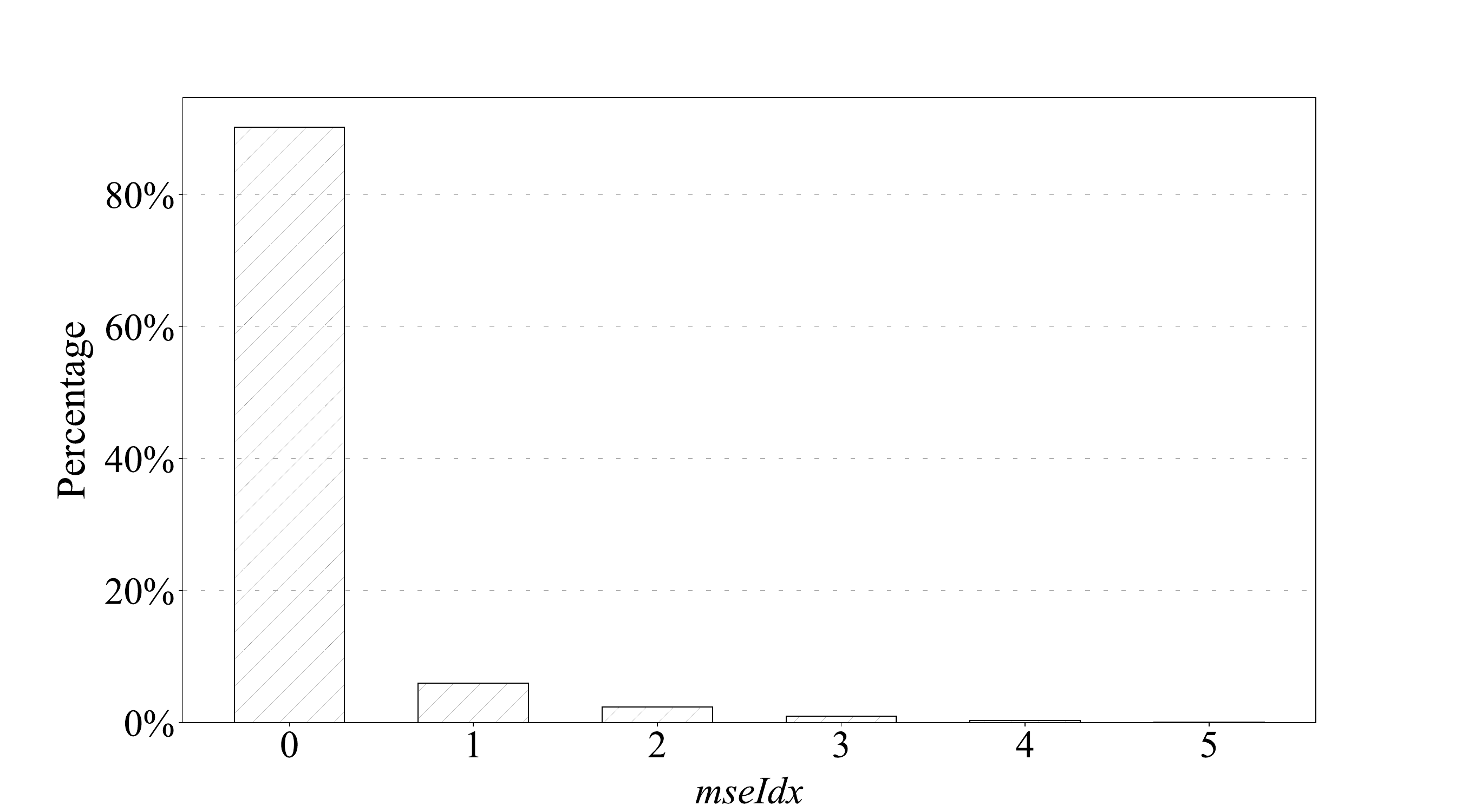}
			\caption{Illustration of the $mseIdx$ distribution. The $mseIdx$ represents the preference of the ultimate encoder-selected MV in the sense of MSE. The smallest $mseIdx$ means that the encoder-selected MV corresponds to the smallest MSE in the MCL, vice versa. The vertical axis represents the percentage of the encoder-selected MV corresponding to different $mseIdx$.} 
			\label{Rect_Mvp2}
		\end{center}
		\vspace{-3mm}
	\end{figure}
	
	\begin{figure}[!t]
		\begin{center}
			\noindent
			\subfigure[\textit{BasketballPass}]{
				\includegraphics[width=1.625in]{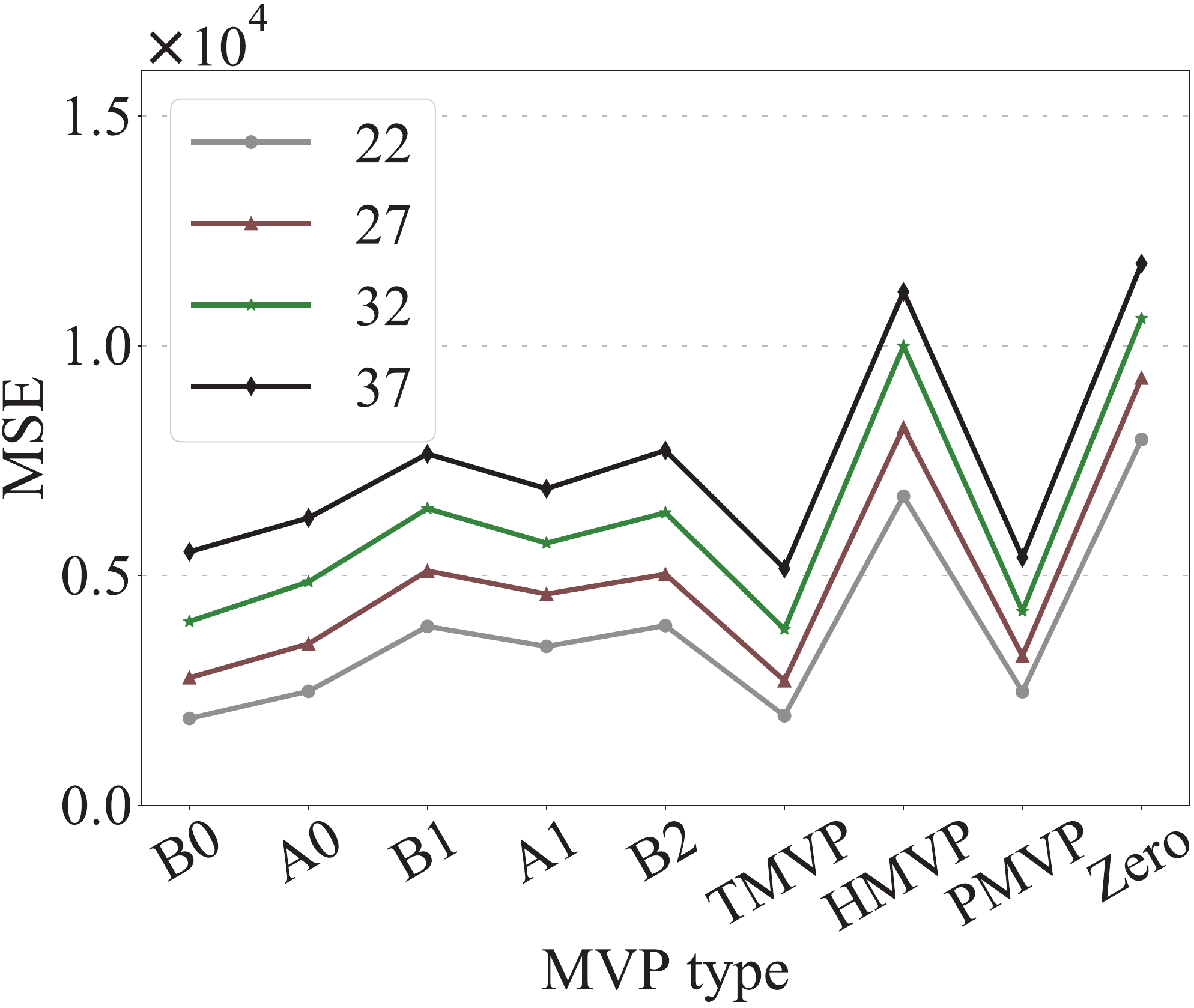}
			}
			\subfigure[\textit{BQSquare}]{
				\includegraphics[width=1.625in]{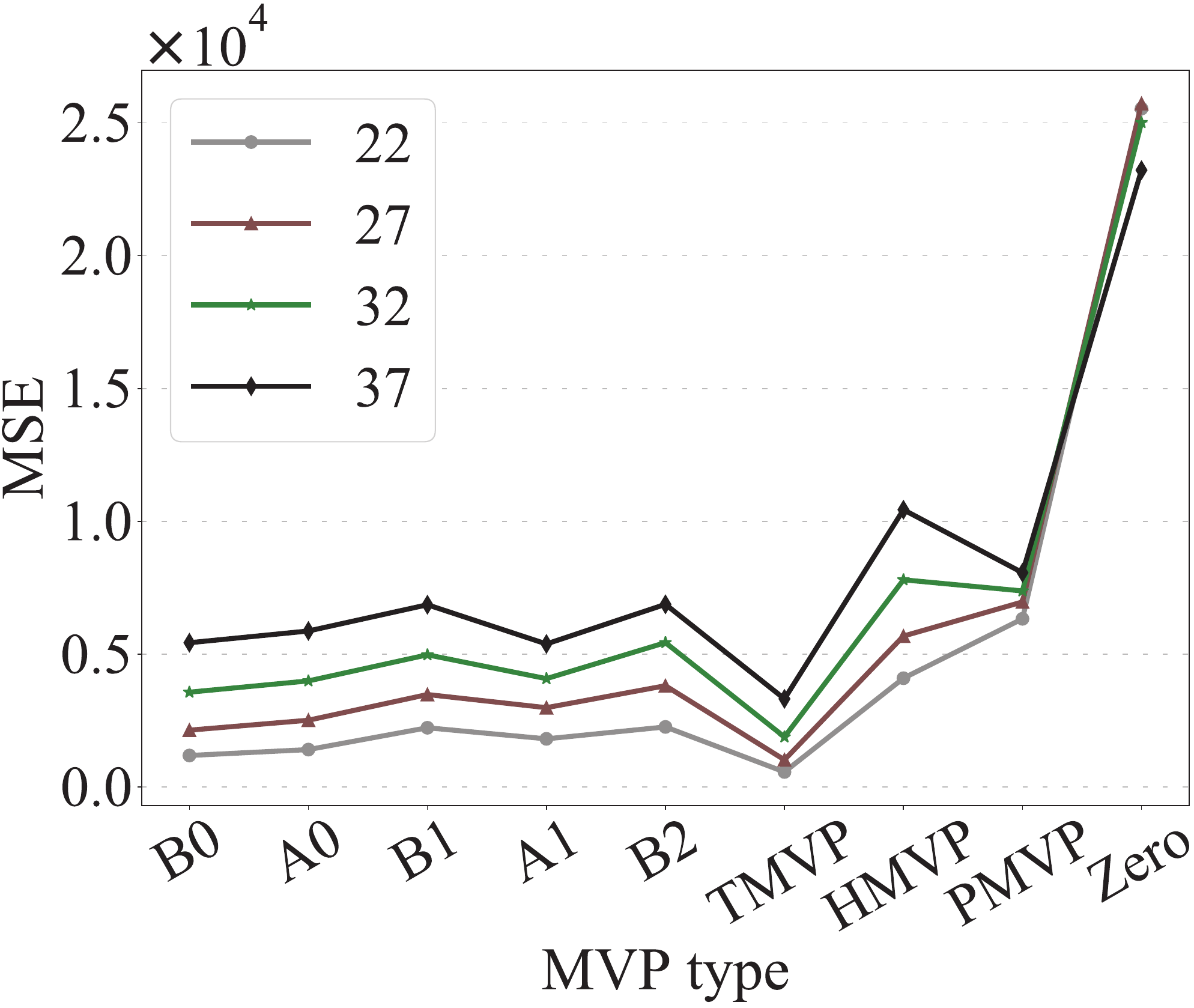}
			}
			\caption{Illustration of the MSE corresponding to different MVPs of rectangular CUs. The horizontal axis represents the nine MVP types, and the vertical axis is the average MSE between the original block and prediction block generated by motion compensation using the corresponding MVP.}
			\label{Rect_Mvp1}
		\end{center}
		\vspace{-3mm}
	\end{figure}
	
	\begin{figure*}[!t]
		\begin{center}
			\noindent
			\subfigure[geoMode=0]{
				\includegraphics[width=1.62in]{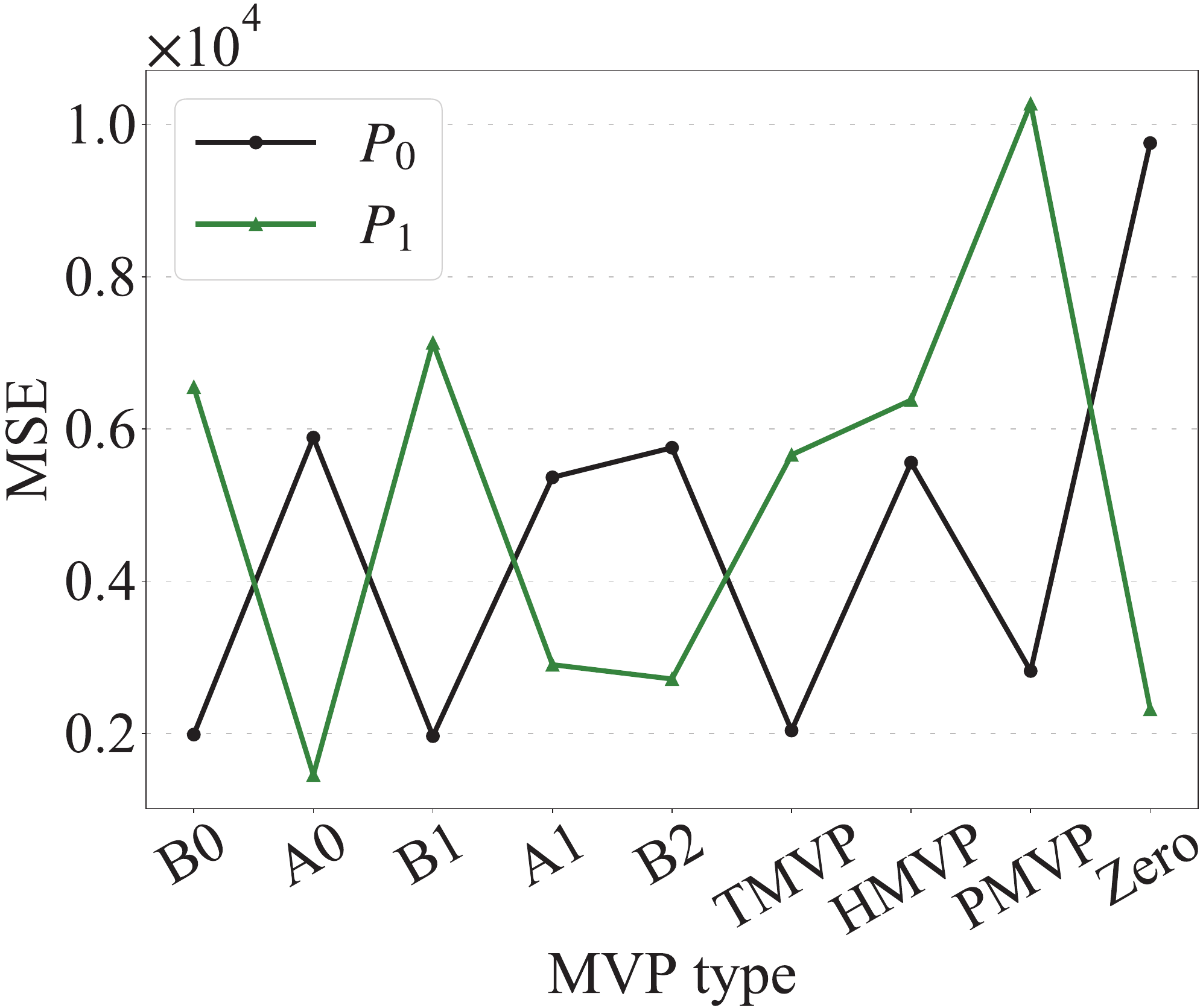}
			}
			\subfigure[geoMode=16]{
				\includegraphics[width=1.62in]{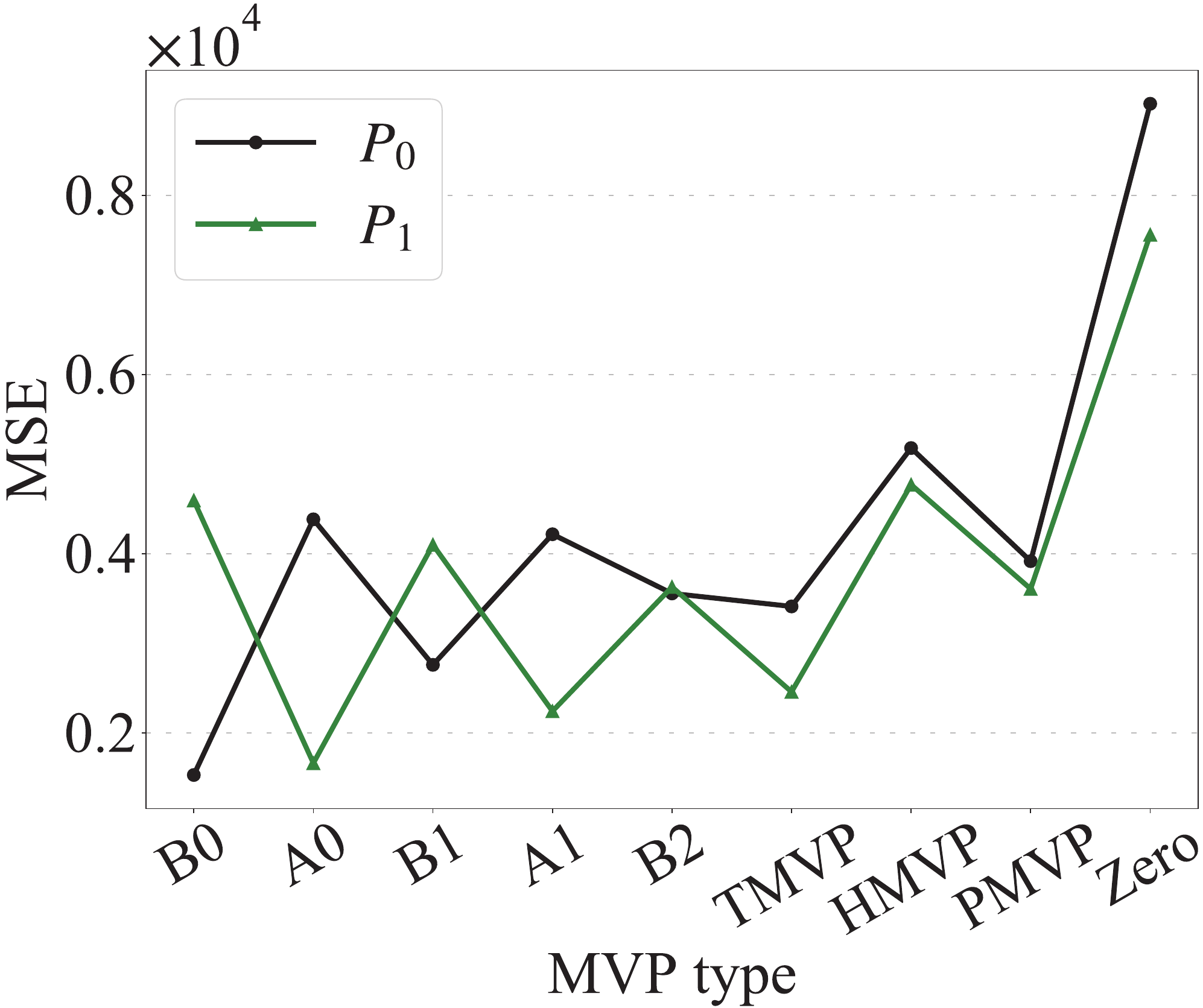}
			}
			\subfigure[geoMode=32]{
				\includegraphics[width=1.62in]{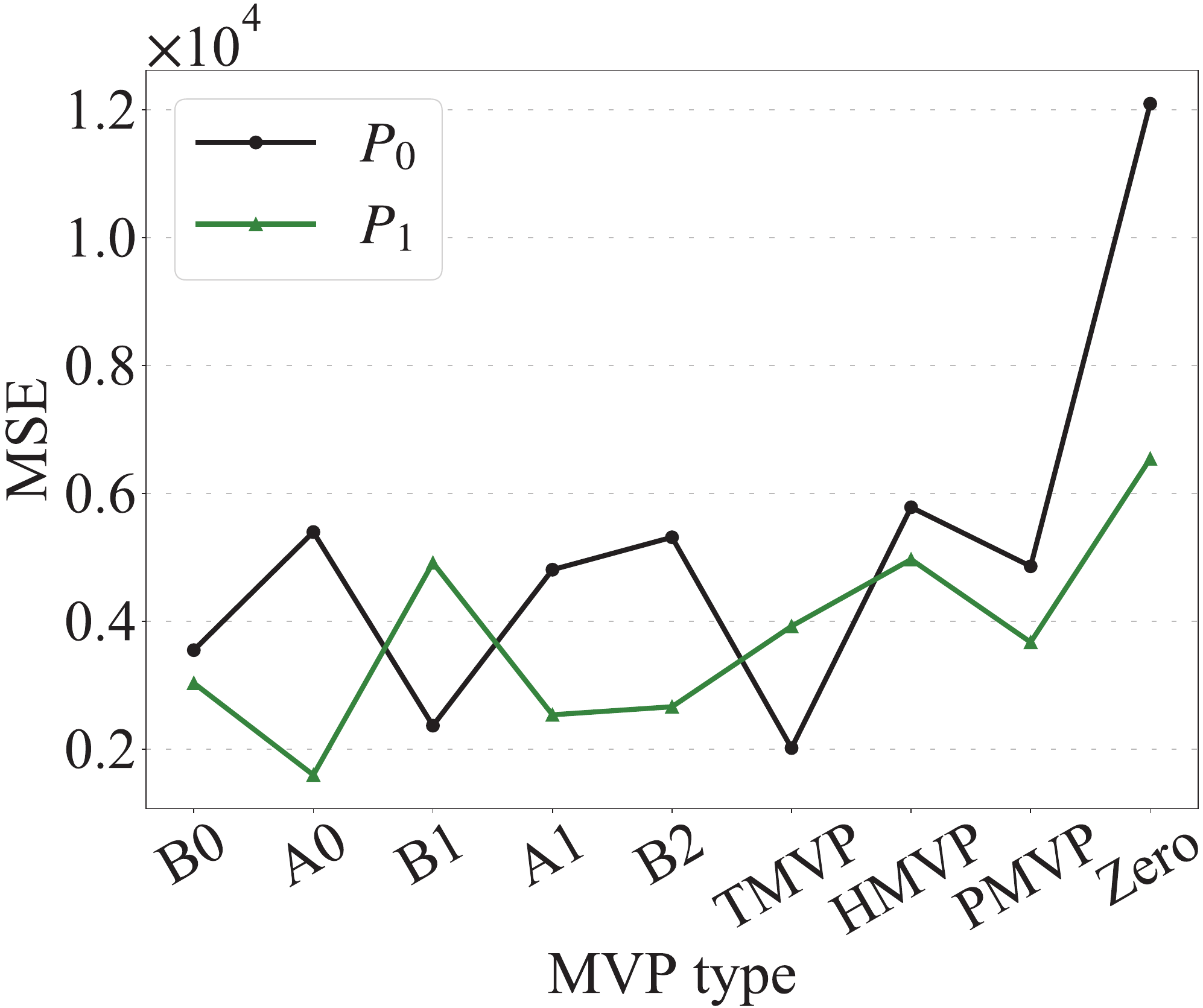}
			}
			\subfigure[geoMode=48]{
				\includegraphics[width=1.62in]{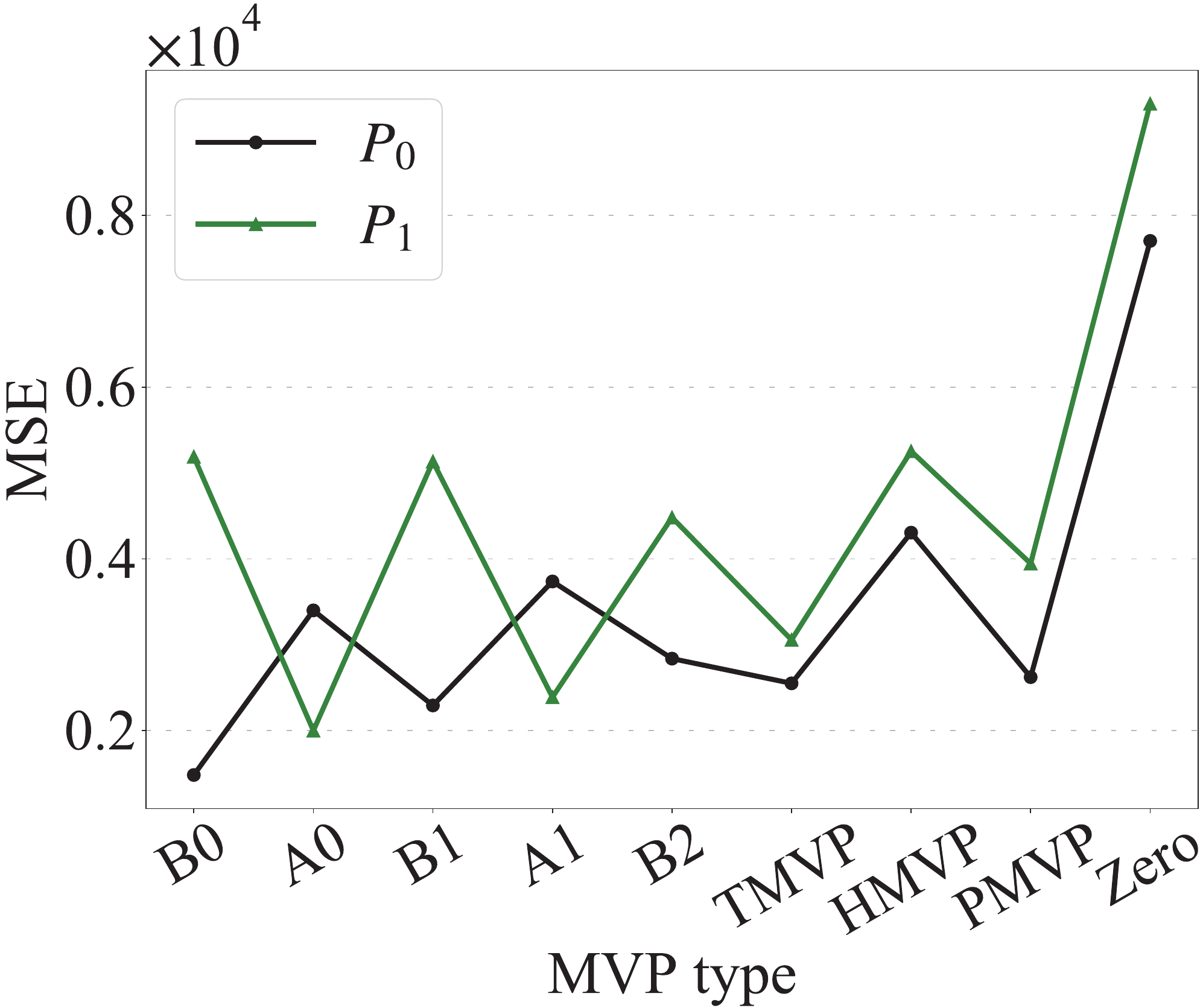}
			}
			\caption{Illustration of the MSE corresponding to different MVPs under certain GEO indices. The horizontal axis represents the nine MVP types, and the vertical axis is the average MSE between the original block and the prediction block generated by the corresponding MVP for each partition.}
			\label{GEO_Mvp1}
		\end{center}
		\vspace{-3mm}
	\end{figure*}
	
	To verify the efficiency of the proposed evaluation criterion, i.e., MSE, we utilize $mseIdx$ to investigate the selection preference of merge candidates in the sense of MSE and RDO. $mseIdx$ is used to represent the preference of the encoder-selected merge candidate in the sense of MSE. The smallest $mseIdx$ indicates that the encoder-selected MV corresponds to the smallest MSE in the MCL $List_{I}$, vice versa. The distribution of $mseIdx$ is shown in Fig.~\ref{Rect_Mvp2}. It is clear that on average 80\% $\sim$ 90\% ultimate MV corresponds to the lowest MSE, which indicates that the prediction block with lower MSE is more likely to be selected by the encoder. Hence, our proposed criterion is efficient for representing the merge candidate preference. 
	
	Furthermore, the MSE distribution of rectangular blocks is analyzed.  First, the CUs coded by regular merge mode~(non-GEO mode) are selected. Then for each selected CU, prediction blocks are generated using each MVP in the MCL. Consequently, MSEs corresponding to the selection preference of each MVP can be calculated. For better illustration, MVPs are classified into nine types, i.e., B0, A0, B1, A1, B2, TMVP, HMVP, PMVP, and Zero. B0, A0, B1, A1 and B2 are corresponding to spatial MVPs with positions ``up'', ``left'', ``up-right'', ``down-left'' and ``up-left.'', respectively~\cite{VTM8.0}. The results of four typical QPs are shown in Fig.~\ref{Rect_Mvp1}. It is obvious that the distribution of MSE is consistent for different QPs and sequences. In particular, the MSEs of the five spatial MVPs show an upward trend with slight fluctuations. TMVP shows similar MSE as the spatial MVPs. An interesting phenomenon lies in that the MSE of HMVP is higher than that of PMVP. The main reason lies in that there are up to 5 HMVPs in a MCL, and the candidates in HMVPs are added to the MCL in descending order of selection probability~\cite{HMVP}. Thus, the MSE variance of prediction blocks generated by HMVP candidates might be larger than other MVPs. Combining the observations from Fig.~\ref{Rect_Mvp2} and Fig.~\ref{Rect_Mvp1}, it can be concluded that conventional rectangular CUs show consistent MVP preference and its MCL construction scheme is based on the selection preference.
	
	\begin{figure}[!t]
		\begin{center}
			\noindent
			\subfigure[geoMode=0]{
				\includegraphics[width=0.7in]{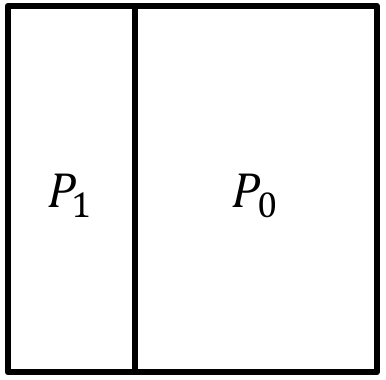}
			}
			\subfigure[geoMode=16]{
				\includegraphics[width=0.7in]{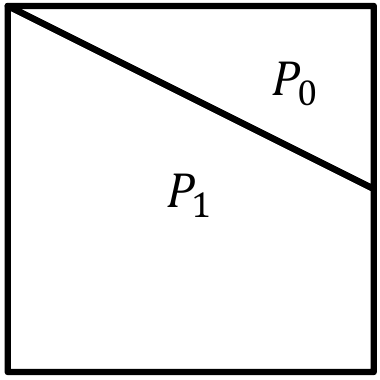}
			}
			\subfigure[geoMode=32]{
				\includegraphics[width=0.7in]{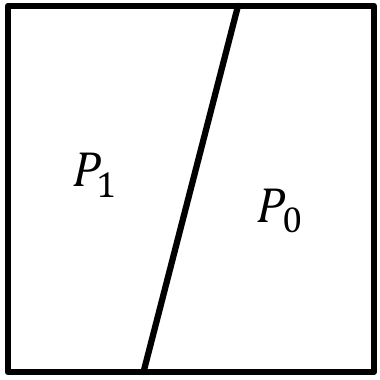}
			}				
			\subfigure[geoMode=48]{
				\includegraphics[width=0.7in]{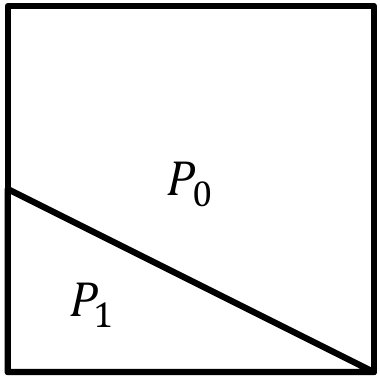}
			}				
			\caption{Illustration of GEO modes selected for the analysis of MVP preference.}
			\label{GEO_part}
		\end{center}
		\vspace{-3mm}
	\end{figure}
	
	In view of this, we further study the relationship between GEO partitioning mode and merge candidate preference based on the MSE of each subpart. In particular, the GEO-coded CUs are collected firstly at the decoder. Then for each collected CU, prediction blocks are generated using the MVs in the MCL. Successively, MSEs corresponding to $P_0$ and $P_1$ are calculated, respectively. For better illustration, only four GEO modes are presented. The selected GEO modes are shown in Fig.~\ref{GEO_part}, and the corresponding MSE results are depicted in Fig.~\ref{GEO_Mvp1}. It is clear that the GEO-coded subparts show more diverse distribution compared to rectangular CUs and the distribution of different GEO modes is also distinct. Taken Fig.~\ref{GEO_Mvp1}(a) as an example, for GEO mode with index 0, ``B0'', ``B1'' and ``TMVP'' are preferred for $P_0$. While, ``A0'' is more relevant to $P_1$. The reason is that $P_0$ is adjacent to the upper block. Hence, its MV is higher correlated with that of the upper blocks and upper-right blocks. 
	Moreover, the MSE between the prediction block generated by TMVP and $P_0$ is relatively lower than the MSE corresponding to TMVP and $P_1$. The main reason might be that TMVP is derived from the collocated block of $P_{0}$ in the reference frame. Furthermore, as the left boundary of $P_1$ is longer than the upper boundary, there is no surprise that the MSE corresponding to the left adjacent block is lower than that of the upper adjacent block. For other GEO modes, the MSE distribution also shows an obvious correlation with partitioning shape. The partitioning shape is determined by the angle $\varphi_i$ and the distance offset $\rho_j$ related to the CU size and $\varphi_i$. Moreover, $i$ and $j$ can be represented by the STGEO mode index. Therefore, the STGEO partitioning shape is related to the CU size and STGEO mode index.  
	
	\begin{table}[t!]
		\begin{center}
			\caption{Syntax structure of the proposed STGEO.} 
			\label{Syntax_pro}
			\begin{tabular}{l|c}
				\thickhline
				\textbf{Syntax}                              & \textbf{Descriptor} \\
				\hline
				STGEO\_parameter\_set~() \{\hspace{0.9cm}    &           \\
				\hspace{0.3cm}\textbf{cu\_stgeo\_flag}       & ae(v)$^*$ \\
				\hspace{0.3cm}if(cu\_stgeo\_flag) \{         &           \\
				\hspace{0.6cm}\textbf{stgeo\_hps\_flag}      & ae(v)     \\
				\hspace{0.6cm}if(stgeo\_hps\_flag)           &           \\
				\hspace{0.9cm}\textbf{stgeo\_hps\_idx}       & ae(v)     \\
				\hspace{0.6cm}else                           &           \\
				\hspace{0.9cm}\textbf{stgeo\_lps\_idx }      & ae(v)     \\
				\hspace{0.6cm}\textbf{merge\_cand\_idx0}     & ae(v)     \\
				\hspace{0.6cm}\textbf{merge\_cand\_idx1}     & ae(v)     \\
				\hspace{0.3cm}\}                             &           \\
				\}                                           &           \\
				\thickhline 
				\multicolumn{2}{c} {$^*$Context-adaptive arithmetic entropy-coded syntax element.} 
			\end{tabular}
		\end{center}
		\vspace{-3mm}
	\end{table}

\begin{table*}[ht!]
	\centering
	\begin{center}
		\caption{Experimental results of the proposed STGEO, Anchor: VTM-8.0 without GEO.} 
		\label{Performance3}
		\setlength{\tabcolsep}{3.2mm}{
			\begin{tabular}{c|c|c c|c c|c c|c c}
				\thickhline
				\hline
				\multirow{2}{*}{\textbf{Class}}&
				\multirow{2}{*}{\textbf{Sequence}}&
				\multicolumn{4}{c|}{\rule{0pt}{8pt} \textbf{GEO}} & \multicolumn{4}{c}{\textbf{STGEO}}\\
				\cline{3-10} & & 
				\textbf{RA}& $\textit{\textbf{UR}}_{\textit{\textbf{GEO}}}$ & \textbf{LDB}&$\textit{\textbf{UR}}_{\textit{\textbf{GEO}}}$& \textbf{RA}&$\textit{\textbf{UR}}_{\textit{\textbf{STGEO}}}$& \textbf{LDB}&$\textit{\textbf{UR}}_{\textit{\textbf{STGEO}}}$\\
				
				\hline
				\multirow{3}{*}{Class A1}
				& \rule{0pt}{8pt} \textit{Tango2}               & -0.59\% & 1.54\%  & -        & -       & -0.78\% & 1.98\%        & -  & -  \\
				& \rule{0pt}{8pt} \textit{FoodMarket4}          & -0.03\% & 0.61\%  & -        & -       & -0.23\% & 1.07\%        & -  & -  \\
				& \rule{0pt}{8pt} \textit{Campfire}             & -0.11\% & 0.97\%  & -        & -       & -0.32\% & 1.42\%        & -  & -  \\
				
				\hline
				\multirow{3}{*}{Class A2}
				& \rule{0pt}{8pt} \textit{CatRobot}             & -0.59\% & 0.98\%  & -        & -       & -0.78\% & 1.45\%        & -  & -  \\
				& \rule{0pt}{8pt} \textit{DaylightRoad2}        & -0.21\% & 1.02\%  & -        & -       & -0.49\% & 1.53\%        & -  & -  \\
				& \rule{0pt}{8pt} \textit{ParkRunning3}         & -0.36\% & 1.98\%  & -        & -       & -0.53\% & 2.46\%        & -  & -  \\
				
				\hline
				\multirow{5}{*}{Class B}
				& \rule{0pt}{8pt} \textit{MarketPlace}          & -0.31\% & 1.02\%  & -1.15\%  & 5.54\%  & -0.51\% & 1.77\%        & -1.28\%  & 6.36\%  \\
				& \rule{0pt}{8pt} \textit{RitualDance}          & -0.75\% & 2.00\%  & -1.24\%  & 5.14\%  & -0.89\% & 2.70\%        & -1.68\%  & 6.08\%  \\
				& \rule{0pt}{8pt} \textit{Cactus}               & -0.53\% & 1.88\%  & -0.96\%  & 4.81\%  & -0.71\% & 2.72\%        & -1.45\%  & 6.15\%  \\
				& \rule{0pt}{8pt} \textit{BasketballDrive}      & -0.23\% & 1.49\%  & -0.73\%  & 4.10\%  & -0.43\% & 2.25\%        & -0.91\%  & 5.05\%  \\
				& \rule{0pt}{8pt} \textit{BQTerrace}            & -0.22\% & 0.84\%  & -0.13\%  & 2.11\%  & -0.40\% & 1.54\%        & -0.38\%  & 2.72\%  \\
				
				\hline
				\multirow{4}{*}{Class C}
				& \rule{0pt}{8pt} \textit{BasketballDrill}      & -0.96\% & 5.37\%  & -1.98\%  & 8.05 \% & -1.36\% & 6.71\%        & -2.74\%  &  9.61\%  \\
				& \rule{0pt}{8pt} \textit{BQMall}               & -3.34\% & 6.56\%  & -3.59\%  & 10.29\% & -3.61\% & 7.69\%        & -4.30\%  & 12.08\%  \\
				& \rule{0pt}{8pt} \textit{Partyscene}           & -0.63\% & 3.27\%  & -1.05\%  & 8.29 \% & -0.92\% & 4.51\%        & -1.49\%  & 10.53\%  \\
				& \rule{0pt}{8pt} \textit{RaceHorsesC}          & -2.01\% & 8.45\%  & -1.92\%  & 11.63\% & -2.35\% & 9.74\%        & -2.43\%  & 13.31\%  \\
				
				\hline
				\multirow{3}{*}{Class E}
				& \rule{0pt}{8pt} \textit{FourPeople}           & -       & -       & -1.30\%  & 3.39\%  & -       & -             & -2.16\%  & 4.45\%  \\
				& \rule{0pt}{8pt} \textit{Johnny}               & -       & -       & -1.85\%  & 2.92\%  & -       & -             & -2.86\%  & 3.67\%  \\
				& \rule{0pt}{8pt} \textit{KristenAndSara}       & -       & -       & -1.28\%  & 3.46\%  & -       & -             & -2.07\%  & 4.52\%  \\
				
				\hline
				\multirow{6}{*}{Average}
				& \rule{0pt}{8pt} {Class A1}                    & -0.24\% & 1.04\%  & -        & -       & -0.44\% & 1.49\%        & -        & -        \\
				& \rule{0pt}{8pt} {Class A2}                    & -0.39\% & 1.33\%  & -        & -       & -0.60\% & 1.81\%        & -        & -        \\
				& \rule{0pt}{8pt} {Class B}                     & -0.41\% & 1.45\%  & -0.84\%  & 4.34\%  & -0.59\% & 2.20\%        & -1.14\%  & 5.27\%   \\
				& \rule{0pt}{8pt} {Class C}                     & -1.74\% & 5.91\%  & -2.13\%  & 9.56\%  & -2.06\% & 7.16\%        & -2.74\%  & 11.38\%  \\
				& \rule{0pt}{8pt} {Class E}                     &       - & -       & -1.48\%  & 3.26\%  & -       & -             & -2.36\%  & 4.21\%   \\
				\hline
				\multicolumn{2}{c|} {\rule{0pt}{8pt}Average}     & -0.72\% & 2.53\%   & -1.43\%  & 5.81\%  & -0.95\%   & 3.30\%   & -1.98\%  & 7.04\%  \\
				\hline
				\multicolumn{2}{c|}{\rule{0pt}{8pt}$\Delta ET$} & \multicolumn{2}{c|}{\rule{0pt}{8pt}103\%} & \multicolumn{2}{c|}{\rule{0pt}{8pt} 104\%} & \multicolumn{2}{c|}{\rule{0pt}{8pt}105\%} & \multicolumn{2}{c}{\rule{0pt}{8pt} 107\%}\\
				\hline
				\multicolumn{2}{c|}{\rule{0pt}{8pt}$\Delta DT$} & \multicolumn{2}{c|}{\rule{0pt}{8pt}100\%} & \multicolumn{2}{c|}{\rule{0pt}{8pt} 100\%} & \multicolumn{2}{c|}{\rule{0pt}{8pt}106\%} & \multicolumn{2}{c}{\rule{0pt}{8pt} 108\%}\\
				
				\thickhline
		\end{tabular}}
	\end{center}
	\vspace{-3mm}
\end{table*}

	Based on the above observations, the merge candidate preference can be predicted using CU size and STGEO mode index. This problem can be formulated as follows. For the subpart $P_*$~($P_0$ or $P_1$) of a CU, given the features, i.e., STGEO mode index $I$ and CU size $S$, we aim to derive the merge candidate selection probability and then infer the MCL. The selection probability can be off-line trained using the MSE. Since the $S$, $I$, and $P_*$ are all discrete features, the selection probability can be mapped into a look-up-table indexed by corresponding feature ($S, I, P_*$) for efficient implementation. Specifically, for a particular ($S, I, P_*$), the MVP type are stored in the look-up table in ascending order of MSE. The inference of MCL is to arrange the {$List_{I}$} by moving the candidates with high selection probabilities to the front.
	More specifically, the number of $S$, $I$ and $P_*$ are 14, 64 and 2, respectively. As ``Zero'' is a default mode to fulfill the MCL, only 8 types of motion candidates are considered, which can be represented by 3 bits. For each feature ($S, I, P_*$), 7 MVP types are stored. Overall, the memory overhead of the look-up table is 4704 bytes.
	
	The syntax design of the proposed scheme is illustrated in Table \ref{Syntax_pro}. In particular, the STGEO serves as a merge mode at the CU level. The on/off flag is signaled by \textit{cu\_stgeo\_flag}. If the STGEO is applied on the current CU, one syntax \textit{stgeo\_hps\_flag} is signaled to indicate whether the selected STGEO mode is in the HPS. If so, the index in the HPS, i.e., \textit{stgeo\_hps\_idx}, is transmitted. Otherwise, \textit{stgeo\_lps\_idx} is signaled. \textit{cu\_stgeo\_flag} is binarized with fixed-length code using one single context model. \textit{stgeo\_hps\_idx} and \textit{stgeo\_lps\_idx} are binarized with truncated binary code and coded using the bypass mode of context-based adaptive binary arithmetic coding~(CABAC). After that, the \textit{merge\_cand\_idx0} and \textit{merge\_cand\_idx1}, indicating the index of the selected merge candidate of $P_0$ and $P_1$, are binarized with 0-th order truncated Rice code and coded using one single context model.
	
    	\begin{table}[ht!]
    	\centering
    	\begin{center}
    		\caption{Experimental results of the proposed STGEO, Anchor: VTM-8.0.} 
    		\label{PerformanceAdded}
    		\setlength{\tabcolsep}{7mm}{
    			\begin{tabular}{c|c |c }
    				\thickhline
    				\hline
    				\multirow{2}{*}{\textbf{Class}}&
    				\multicolumn{2}{c}{\rule{0pt}{8pt} \textbf{BDRate-Y}}\\
    				\cmidrule{2-3} &
    				\textbf{RA}& \textbf{LDB} \\
    				\midrule
    				Class A1   & -0.20\%    & -                \\
    				Class A2   & -0.21\%    & -                 \\
    				Class B    & -0.19\%    & -0.31\%   \\
    				Class C    & -0.32\%    & -0.61\%  \\
    				Class E    &       -    & -0.88\%  \\
    				\midrule
    				Average & -0.23\%   & -0.55\%  \\
    				\midrule
    				$\Delta ET$ & \multicolumn{1}{c|}{\rule{0pt}{8pt}102\%} & \multicolumn{1}{c}{\rule{0pt}{8pt} 102\%}\\
    				\midrule
    				$\Delta DT$ & \multicolumn{1}{c|}{\rule{0pt}{8pt}106\%} & \multicolumn{1}{c}{\rule{0pt}{8pt} 107\%}\\
    				\thickhline
    		\end{tabular}}
    	\end{center}
    	\vspace{-3mm}
    \end{table}

	\section{Experimental Results}\label{experimental}
	To evaluate the effectiveness of the proposed method, we integrate it into the VVC reference software VTM-8.0. The videos recommended by JVET are involved in the experiment. Two configurations, RA and LDB, conforming to the common test condition~\cite{CTC} are used in the simulation. The QPs are set as \{22, 27, 32, 37\}. The coding performance is measured by Bjontegaard's method~\cite{BDrate} in terms of BD-rate, and negative BD-rate indicates the performance gain.
	
		\begin{table*}[ht!]
		\centering
		\begin{center}
			\caption{Experimental results of the proposed STGEO without \textit{T2} module~(Anchor: VTM-8.0 without GEO).} 
			\label{Performance1}
			\setlength{\tabcolsep}{3.5mm}{
				\begin{tabular}{c|c c c c c|c c c c c}
					\thickhline
					\hline
					\multirow{2}{*}{\textbf{Class}}&
					\multicolumn{5}{c|}{\rule{0pt}{8pt} \textbf{RA}} & \multicolumn{5}{c}{\textbf{LDB}}\\
					\cmidrule{2-11} &
					\textbf{Y}& \textbf{U}& \textbf{V}& $\textit{\textbf{MR}}$&  $\textit{\textbf{UR}}_{\textit{\textbf{T1}}}$ & \textbf{Y}& \textbf{U}& \textbf{V}& $\textit{\textbf{MR}}$&  $\textit{\textbf{UR}}_{\textit{\textbf{T1}}}$\\
					
					\midrule
					Class A1   & -0.35\% & -0.73\%  & -0.47\%  & 59.5\%  & 1.26\% & -       & -        & -        & -     & -  \\
					Class A2   & -0.57\% & -0.68\%  & -0.83\%  & 60.2\%  & 1.55\% & -       & -        & -        & -     & -  \\
					Class B    & -0.56\% & -0.88\%  & -0.66\%  & 60.0\%  & 1.82\% & -0.98\% & -1.32\%  & -0.95\%  & 59.1\%  & 4.70\% \\
					Class C    & -1.96\% & -2.54\%  & -2.95\%  & 58.3\%  & 6.79\% & -2.45\% & -3.17\%  & -2.65\%  & 56.6\%  & 10.15\%\\
					Class E    &       - & -        &  -       & -       & -      & -1.97\% & -0.79\%  & -1.13\%  & 59.9\%  & 3.50\% \\
					\midrule
					Average & -0.89\%  & -1.25\%  & -1.26\% & 59.5\% & 2.98\% & -1.72\%  & -1.80\% & -1.56\%  & 58.4\% & 6.22\% \\
					\midrule
					$\Delta ET$ & \multicolumn{5}{c|}{\rule{0pt}{8pt}105\%} & \multicolumn{5}{c}{\rule{0pt}{8pt} 107\%}\\
					\midrule
					$\Delta DT$ & \multicolumn{5}{c|}{\rule{0pt}{8pt}105\%} & \multicolumn{5}{c}{\rule{0pt}{8pt} 108\%}\\
					\thickhline
			\end{tabular}}
		\end{center}
		\vspace{-3mm}
	\end{table*}
	
	\begin{table*}[ht!]
		\centering
		\begin{center}
			\caption{Experimental results of the proposed STGEO without \textit{T1} module~(Anchor: VTM-8.0 without GEO).} \label{Performance2}
			\setlength{\tabcolsep}{3.7mm}{
				\begin{tabular}{c|c c c c c|c c c c c}
					\thickhline
					\hline
					\multirow{2}{*}{\textbf{Class}}&
					\multicolumn{5}{c|}{\rule{0pt}{8pt} \textbf{RA}} & \multicolumn{5}{c}{\textbf{LDB}}\\
					\cmidrule{2-11} &
					\textbf{Y}&  \textbf{U}& \textbf{V}&  $\textit{\textbf{BS}}$&  $\textit{\textbf{UR}}_{\textit{\textbf{T2}}}$&   \textbf{Y}&  \textbf{U}& \textbf{V}&  $\textit{\textbf{BS}}$&  $\textit{\textbf{UR}}_{\textit{\textbf{T2}}}$\\
					
					\midrule
					Class A1   & -0.30\% & -0.45\%  & -0.50\%  & 0.13  & 1.22\% & -       & -        & -        & -     & -  \\
					Class A2   & -0.52\% & -0.60\%  & -0.87\%  & 0.09  & 1.50\% & -       & -        & -        & -     & -  \\
					Class B    & -0.45\% & -0.94\%  & -0.76\%  & 0.08  & 1.66\% & -1.00\% & -1.52\%  & -1.15\%  & 0.08  & 4.91\% \\
					Class C    & -1.93\% & -2.85\%  & -2.74\%  & 0.09  & 6.53\% & -2.52\% & -3.66\%  & -2.97\%  & 0.11  & 10.80\%\\
					Class E    &       - & -        &  -       & -     & -      & -1.98\% & -1.04\%  & -0.79\%  & 0.07  & 3.95\% \\
					\midrule
					Average    & -0.83\%  & -1.28\%  & -1.26\% & 0.09  & 2.84\% & -1.75\% & -2.11\% & -1.67\%  & 0.09 & 6.63\% \\
					\midrule
					$\Delta ET$ & \multicolumn{5}{c|}{\rule{0pt}{8pt}104\%} & \multicolumn{5}{c}{\rule{0pt}{8pt} 105\%}\\
					\midrule
					$\Delta DT$ & \multicolumn{5}{c|}{\rule{0pt}{8pt}101\%} & \multicolumn{5}{c}{\rule{0pt}{8pt} 101\%}\\
					\thickhline
			\end{tabular}}
		\end{center}
		\vspace{-3mm}
	\end{table*}

	Table \ref{Performance3} shows the coding performance of the proposed STGEO scheme on luma component compared to the VTM-8.0 without GEO for each sequence under RA and LDB. Meanwhile, the encoding and decoding complexity is calculated by,
	\begin{equation}
	\Delta ET = \frac{TEnc_{\mathrm{test}}}{TEnc_\mathrm{{anchor}}} \times 100\%,
	\end{equation}

	\begin{equation}
	\Delta DT = \frac{TDec_{\mathrm{test}}}{TDec_{\mathrm{anchor}}} \times 100\%,
	\end{equation}
	where $TEnc_{\mathrm{test}}$ and $TEnc_\mathrm{{anchor}}$ represent the consumed encoding time with the tested scheme and the anchor, respectively. Analogously, $TDec_{\mathrm{test}}$ and $TDec_{\mathrm{anchor}}$ denote the decoding time of the tested scheme and anchor, respectively. To further explore the efficiency of the proposed method, we conduct statistical experiments on all test sequences with respect to the STGEO usage ratio at the decoder. In particular, we analyze the percentage of STGEO coded area, which is formulated as follows,
	\begin{equation}
	UR_{*} = \frac{N_{\mathrm{test}}}{N_{\mathrm{total}}} \times 100\%,
	\end{equation}
	where $N_{\mathrm{test}}$ is the number of pixels coded by STGEO or GEO, and $N_{\mathrm{total}}$ represents the number of pixels in B-frames of the test sequences, as STGEO is only applied on B-frames.
	
	It is observed that the proposed STGEO scheme achieves 0.95\% and 1.98\% BD-rate savings under RA and LDB configurations on average, which outperforms the existing GEO. The highest coding gain is 3.61\% and 4.30\% for \textit{BQMall} under RA and LDB configurations, respectively. The main reason is that this sequence contains sharp and clear boundaries of moving objects. As such, the STGEO can be used predominantly for the coding of object boundaries. Compared to the existing GEO, the proposed STGEO performs relatively stable for all sequences with a minimum of 0.13\% coding gain for \textit{MarketPlace} and a maximum of 1.01\% coding gain for \textit{Johnny} under LDB configuration. The usage ratio of the proposed STGEO $UR_{STGEO}$ also shows a stable increase compared to $UR_{GEO}$ with a minimum of 0.61\% increment for \textit{BQTerrance} and a maximum of 2.24\% increment for \textit{Partyscene} under LDB configuration. It is noted that the coding performances of STGEO in Class E are notably better than that of GEO under LDB configuration, as shown in Table \ref{PerformanceAdded}. The particular favorable results of these sequences are mainly attributable to the fact that the sequences of class E are conference videos with stable motion and clear boundaries, which own more predictable coding modes. In addition, coding performance is better for videos with small resolutions. The reason is that videos with smaller resolutions tend to have more abundant texture and structure information in given block size. The second column in Fig.~\ref{edge_back_residual} can better shows this phenomenon, in which the percentage of pixels in edge area is higher for \textit{BQMall}~(832$\times$480) compared to \textit{Cactus}~(1920$\times$1080).
	
	Furthermore, the performance of STGEO for LDB is generally better than that for RA, which behaves similarly to GEO. The $UR_*$ indicates that STGEO and GEO are more often selected in LDB than RA, which yields the higher coding gain. The reason is that the more flexible reference picture structure of RA, including both forward and backward reference frames, leads to more accurate block-based motion compensation. Hence, the performance room for the STGEO mode is lower under the RA configuration. 
	
	Since STGEO contains two non-overlapped modules, \textit{most probable STGEO mode prediction} \textit{T1} and \textit{probability-based MCL inference} \textit{T2}, the performances of the two modules are evaluated separately, as shown in Table \ref{Performance1} and Table \ref{Performance2}. Regarding the \textit{T1} module, we use the HPS hit ratio at the decoder to validate the efficiency of mode prediction, which is formulated as follows,  
	\begin{equation}
	MR = \frac{N_{hps}}{N_{STGEO}} \times 100\%,
	\end{equation}
	where $N_{hps}$ is the number of CUs with STGEO mode in the HPS, and $N_{STGEO}$ represents the number of STGEO-coded CUs. As for the \textit{T2} module, we evaluate the effectiveness by bit saving, which is calculated by,
	\begin{equation}
	BS = \frac{bit_{org} - bit_{pro}}{N_{STGEO}},
	\end{equation}
	where $bit_{org}$ and $bit_{pro}$ represent the number of bits consumed by the coding of STGEO merge indices using the original MCL and the proposed inferred MCL at the decoder, respectively. Combining Table \ref{Performance3}, Table \ref{Performance1} and Table \ref{Performance2}, it can be obviously seen that the BD-rate improvements provided by the two modules individually are addable and can together form a better R-D performance. As for \textit{T1} and \textit{T2}, they both show stable coding gains under RA and LDB configurations. The $MR$ in Table \ref{Performance1} provides useful evidence that our HPS construction scheme can efficiently predict the high-preference STGEO modes. Besides, higher $MR$ can be observed for Class E, which is in accordance with the coding performance improvements. Moreover, the $BS$ in Table \ref{Performance2} shows that 0.09 and 0.09 bits per STGEO-coded CU are saved for RA and LDB, respectively. The coding gains of LDB configuration are comparable for the two modules, which are 1.72\% and 1.75\% on average for \textit{T1} and \textit{T2}, respectively. While, the coding performance of \textit{T1} is 0.06\% higher than that of \textit{T2} under RA configuration. The two modules both show relative high coding gain for the conference videos, i.e., \textit{FourPeople}, \textit{Johnny} and \textit{KristenAndSara}. 
		
	Concerning the computational complexity, on average 5\% and 7\% encoding time increment of STGEO is observed for RA and LDB configurations, respectively. The time increment is negligible compared to the conventional GEO. The decoding time is increased by 6\% and 8\% on average, which is mainly introduced by the module \textit{T1} which has 5\% and 8\% decoding time increment.

	\section{Conclusion}\label{conclusion}
	In this paper, a spatio-temporal correlation guided geometric partitioning scheme was proposed to improve the coding efficiency. We first identified the problem in the existing GEO: a heavy burden of side information signaling, including the partitioning mode and motion information. Then we proposed an adaptive mode prediction and coding method to reduce the bits consumed for the representation of the above-mentioned side information. In particular, the STGEO mode set is split into two subsets, HPS and LPS. The modes in the HPS have relatively high selection probabilities and can be represented by smaller indices. As such, the STGEO mode can be represented using fewer bits on average. As for the motion information, it is represented by the index in MCL, which is adaptively inferred based on the off-line trained selection probability. Specifically, the motion candidates with higher selection probabilities are put at the front location of the MCL, vice versa. Experimental results verify that STGEO is effective for side information representation and can outperform the existing GEO in VTM-8.0. Compared to VTM-8.0 without GEO, on average 0.95\% and 1.98\% bit-rate savings can be achieved by the proposed method for RA and LDB configurations, respectively. 
	
	The analysis and observation in this paper provide more directions for further improving the representation capability of block partitioning, such as ternary tree geometric partitioning, interweaved partitioning, and so on. The motion information for each subpart can be selected from motion candidates of extended areas, including adjacent and non-adjacent areas. For these methods, the side information representation and signaling method for partitioning mode and motion information will be more important.
	
	\section*{Acknowledgment}
	The authors would like to thank the associate editor and anonymous reviewers for their valuable comments that significantly helped them in improving the quality of the paper.
	
	\bibliographystyle{IEEEtran}
	\bibliography{icme2020template}
	
		\begin{IEEEbiography}[{\includegraphics[width=1in,height=1.25in,clip,keepaspectratio]{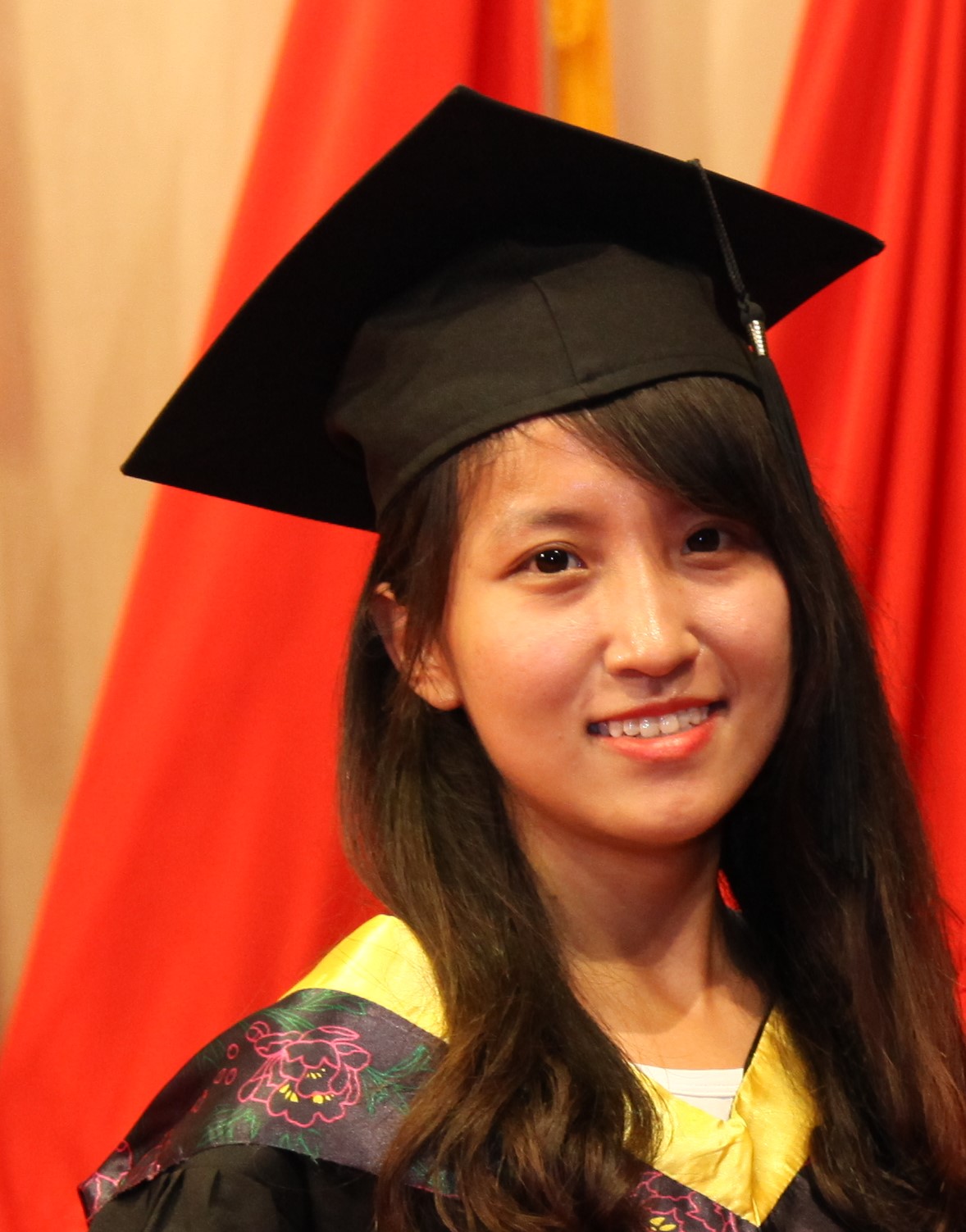}}]{Xuewei Meng}
			received B.E. degree in communication engineering from Beijing University of Posts and Telecommunications, Beijing, China, in 2017. She is currently pursuing the Ph.D. degree with the Department of Computer Science in Peking University, Beijing, China. Her research interests include video compression and video coding standard. She is actively participating in the research of \textit{Versatile Video Coding}~(VVC) standard.
		\end{IEEEbiography}
	
		\begin{IEEEbiography}[{\includegraphics[width=1in,height=1.25in,clip,keepaspectratio]{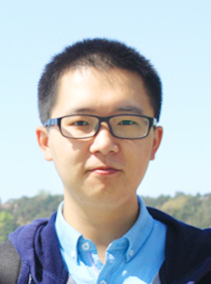}}]{Chuanmin Jia}
			received B.E. degree in computer science from Beijing University of Posts and Telecommunications, Beijing, China, in 2015 and the Ph.D. degree in computer application technology from Peking University, Beijing, China, in 2020. He was a visiting student with Video Lab, New York University, NY, USA, in 2018. He is currently working as Boya postdoc fellow with the Department of Computer Science, Peking University, Beijing, China. His research interests include image/video compression, multimedia signal processing and analysis.
			
			He is the recipient of Best Paper Award of Pacific-Rim Conference on Multimedia in 2017, Best Student Paper Award of IEEE International Conference on Multimedia Information Processing and Retrieval 2019, and also the co-recipient of Best Paper Award of IEEE Multimedia Magazine in 2018.
		\end{IEEEbiography}
	
		\begin{IEEEbiography}[{\includegraphics[width=1in,height=1.25in,clip,keepaspectratio]{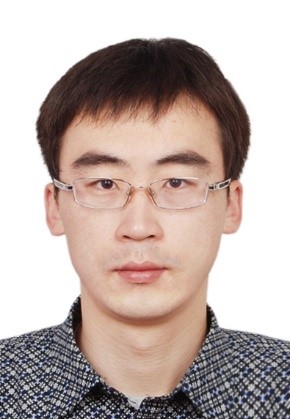}}]{Xinfeng Zhang} (M'16-SM'20) received the B.S. degree in computer science from the Hebei University of Technology, Tianjin, China, in 2007, and the Ph.D. degree in computer science from the Institute of Computing Technology, Chinese Academy of Sciences, Beijing, China, in 2014. From 2014 to 2017, he was a Research Fellow with the Rapid-Rich Object Search Lab, Nanyang Technological University, Singapore. From Oct. 2017 to Oct. 2018, he was a Post-Doctoral Fellow with the School of Electrical Engineering System, University of Southern California, Los Angeles, CA, USA. From Dec. 2018 to Aug. 2019, he was a Research Fellow with the department of Computer Science, City University of Hong Kong.
			
			He currently is an Assistant Professor with the School of Computer Science and Technology, University of Chinese Academy of Sciences. He authored more than 150 refereed journal/conference papers and received the Best Paper Award of IEEE Multimedia 2018, the Best Paper Award at the 2017 Pacific-Rim Conference on Multimedia (PCM) and the Best Student Paper Award in IEEE International Conference on Image Processing 2018. His research interests include video compression and processing, image/video quality assessment, and 3D point cloud processing.
			
		\end{IEEEbiography}

		\begin{IEEEbiography}[{\includegraphics[width=1in,height=1.25in,clip,keepaspectratio]{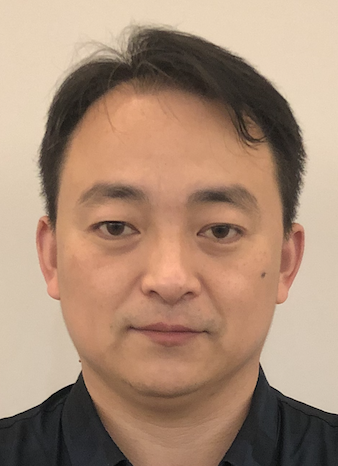}}]{Shanshe Wang}
			received the B.S. degree from the Department of Mathematics, Heilongjiang University, Harbin, China, in 2004, the M.S. degree in computer software and theory from Northeast Petroleum University, Daqing, China, in 2010, and the Ph.D. degree in computer science from the Harbin Institute of Technology. He held a postdoctoral position with Peking University from 2016 to 2018. He joined the School of Electronics Engineering and Computer Science, Institute of Digital Media, Peking University, Beijing, where he is currently a associated researcher. His current research interests include video compression and image and video quality assessment.
		\end{IEEEbiography}
	
		\begin{IEEEbiography}[{\includegraphics[width=1in,height=1.25in,clip,keepaspectratio]{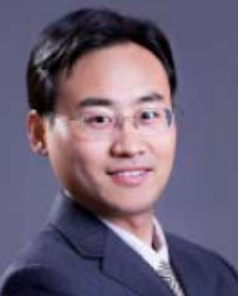}}]{Siwei Ma} (Senior Member, IEEE) received the B.S. degree from Shandong Normal University, Jinan, China, in 1999, and the Ph.D. degree in computer science from the Institute of Computing Technology, Chinese Academy of Sciences, Beijing, China, in 2005. He held a postdoctoral position with the University of Southern California, Los Angeles, CA, USA, from 2005 to 2007. He joined the School of Electronics Engineering and Computer Science, Institute of Digital Media, Peking University, Beijing, where he is currently a Professor. He has authored over 300 technical articles in refereed journals and proceedings in image and video coding, video processing, video streaming, and transmission. He served/serves as an Associate Editor for the IEEE Transactions on Circuits and Systems for Video Technology and the Journal of Visual Communication and Image Representation.
		\end{IEEEbiography}
	
	
	
	
	
\end{document}